\documentclass[journal]{IEEEtran}
\usepackage[left=0.75in, right=0.75in, top=0.75in, bottom=0.75in]{geometry}
\usepackage[utf8]{inputenc}
\usepackage[T1]{fontenc}
\usepackage{xcolor}
\usepackage{cite}
\usepackage{float}
\usepackage{hyperref}
\usepackage[utf8]{inputenc} 
\usepackage[T1]{fontenc}    
\usepackage{hyperref}       
\usepackage{url}            
\usepackage{booktabs}       
\usepackage{amsfonts}       
\usepackage{nicefrac}       
\usepackage{microtype}      
\usepackage{bbold}
\usepackage{capt-of}
\usepackage{mathtools}
\usepackage{fancyvrb}
\usepackage[ruled,vlined]{algorithm2e}
\usepackage{algpseudocode}

\usepackage{amsmath, amsthm, amssymb, graphicx, cite, epsfig,epstopdf,color,soul}
\usepackage{tabularx}

\allowdisplaybreaks
\usepackage{color}
\usepackage{enumerate}
\usepackage[shortlabels]{enumitem}
\usepackage{multicol}
\usepackage{empheq}
\newcommand{\aqeel}[2]{%
    \ifthenelse{\equal{#1}{error}}{{\color{red} [\textbf{aqeel}: #2]}}{%
    \ifthenelse{\equal{#1}{warn}}{{\color{blue} [\textbf{aqeel}: #2]}}{}}}
\usepackage{booktabs}
\usepackage{amsmath,amssymb,amsfonts, lipsum}
\usepackage{graphicx}
\usepackage{textcomp}
\usepackage{xcolor}
\usepackage{optidef}

\usepackage[font=small,labelfont=bf,justification=justified,format=plain]{caption}

\begin{document}
\title{Multi-Task Federated Reinforcement Learning with Adversaries}
\author{\IEEEauthorblockN{Aqeel Anwar\textsuperscript{1}, Arijit Raychowdhury\textsuperscript{2}} \\
\IEEEauthorblockA{\textit{Department of Electrical and Computer Engineering} \\
\textit{Georgia Institute of Technology, Atlanta, GA, USA}\\
\textit{aqeel.anwar@gatech.edu\textsuperscript{1}, arijit.raychowdhury@ece.gatech.edu\textsuperscript{2}}}
}

\maketitle
\thispagestyle{empty}
\pagestyle{plain}

\begin{abstract}
Reinforcement learning algorithms, just like any other Machine learning algorithm pose a serious threat from adversaries. The adversaries can manipulate the learning algorithm resulting in non-optimal policies. In this paper, we analyze the Multi-task Federated Reinforcement Learning algorithms, where multiple collaborative agents in various environments are trying to maximize the sum of discounted return, in the presence of adversarial agents. We argue that the common attack methods are not guaranteed to carry out a successful attack on Multi-task Federated Reinforcement Learning and propose an adaptive attack method with better attack performance. Furthermore, we modify the conventional federated reinforcement learning algorithm to address the issue of adversaries that works equally well with and without the adversaries. Experimentation on different small to mid-size reinforcement learning problems show that the proposed attack method outperforms other general attack methods and the proposed modification to federated reinforcement learning algorithm was able to achieve near-optimal policies in the presence of adversarial agents.
\end{abstract}

\flushbottom
\maketitle
\thispagestyle{empty}

\section{Introduction}
In the past decade, Reinforcement Learning (RL) has gained wide popularity in solving complex problems in an online fashion for various problem sets such as game playing \cite{mnih2015human}, autonomous navigation \cite{anwar2018navren, wang2019autonomous}, robotics \cite{kober2013reinforcement} and network security \cite{xiao2018security}. In most of the real-life cases where we do not have complete access to the system dynamics, conventional control theory fails to provide optimum solutions. Model-free RL on the other hand uses heuristics to explore and exploit the environment to achieve the underlying goal. With a boom in Internet of Things (IoT) devices \cite{li2015internet}, we have a lot of compute power at our disposal. The problem, however, is that the compute is distributed. Distributed algorithms have been studied to take advantage of these distributed compute agents. Conventional method consist of using these IoT as data-collectors and then using a centralized server to train a network on the collected data. Federated Learning, introduced by Google \cite{bonawitz2019towards, bonawitz2017practical, konevcny2016federated} is a distributed approach to machine learning tasks enabling model training on large sets of decentralized data by individual agents. The key idea behind federated learning is to preserve the privacy of the data to the local node responsible for generating it. The training data is assumed to be local only, the agents however, can share the model parameter that is learned. This model sharing serves two purpose. Primarily it ensures the privacy of the data being generated locally, secondly in some of the cases the size of the model parameter might be much smaller than the size of the local data, hence sharing the model parameter instead of the data might save up on the communication cost involved. Federated learning has also been considered in the context of Reinforcement learning problem for both multi-agent RL \cite{kumar2017federated, zhuo2019federated, palmer2017lenient, hernandez2019survey} and multi-task RL \cite{lim2020federated, liu2019lifelong, zeng2020decentralized} where multiple RL agents either in a single or multiple environments try to jointly maximize the collective or sum of individual discounted returns, respectively.
\begin{figure}[t]
  \includegraphics[width=\linewidth]{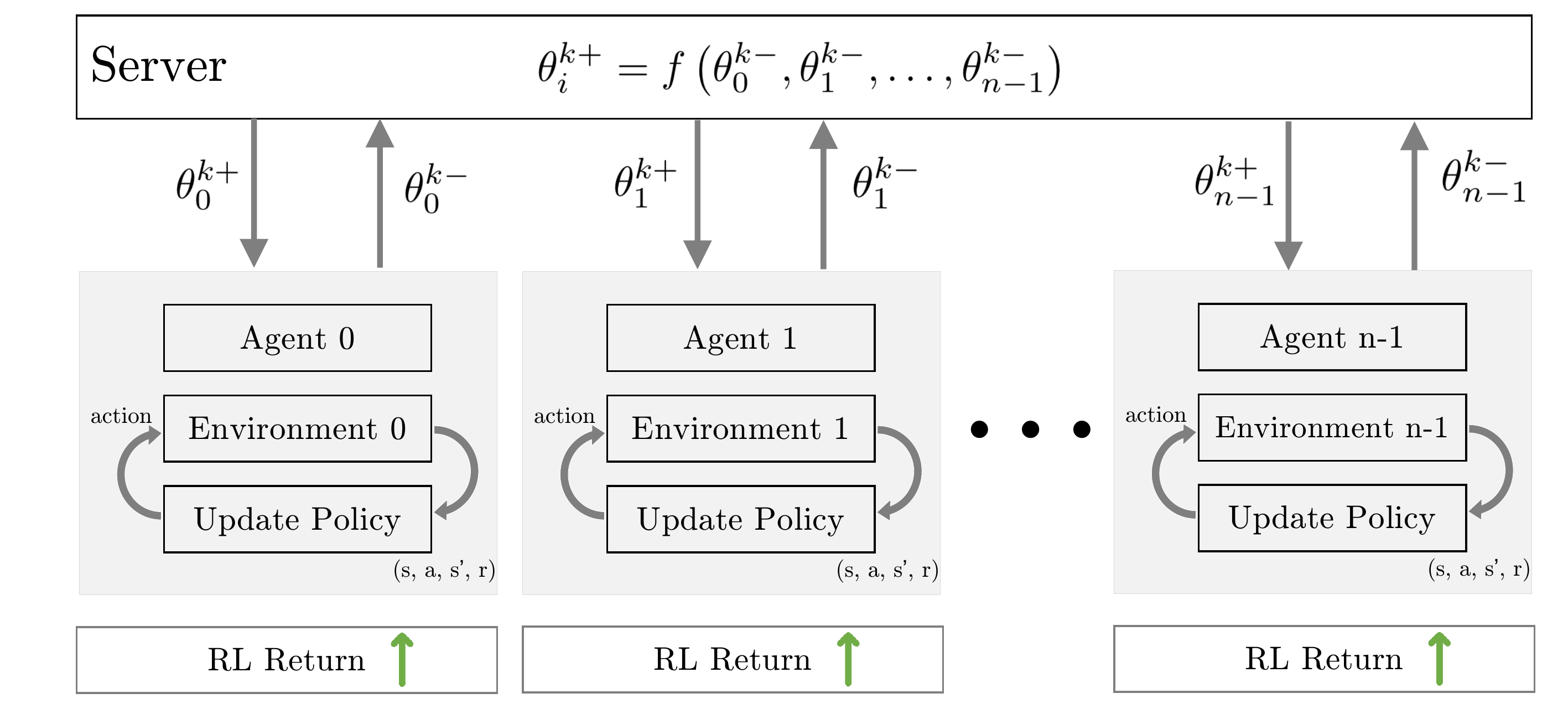}
  \caption{Federated RL - The idea is to learn a common unified policy without sharing the local training data that works good enough for all the environments}
  \label{fig:block}
\end{figure}

While ML algorithms have proven to provide superior accuracy over conventional methods, they pose a threat from adversarial manipulations. Adversaries can use a variety of attack models to manipulate the model either in the training or the inference phase leading to decreased accuracy or poor policies. Common attack methods include data-poisoning and model poisoning where the adversary tries to manipulate the input data or directly the learned model respectively. In this paper, we propose and analyze a model-poisoning attack for the Multi-task Federated RL (MT-FedRL) problem and modify the conventional Federated RL approach to provide protection from model poisoning attacks.

The contributions of this paper are as follows

\begin{itemize}
    \item We carry out a detailed study on the multi-task federated RL (MT-FedRL) with model-poisoning adversaries on medium and large size problems of grid-world (GridWorld) and drone autonomous navigation(AutoNav).
    \item We argue that the general adversarial methods are not good enough to create an effective attack on MT-FedRL, and propose a model-poisoning attack methodology \textit{AdAMInG} based on minimizing the information gain during the MT-FedRL training.
    \item Finally, we address the adversarial attack issue by proposing a modification to the general FedRL algorithm, \texttt{ComA-FedRL}, that works equally well with and without adversaries.
\end{itemize}

The rest of the paper is organized as follows. In Sec. \ref{sec:related_work} we provide a background on the related work in the area of adversarial machine learning. Sec. \ref{sec:mtrl} and \ref{sec:mtrl_adv}formally defines the MT-FedRL problem and the adversarial formulation respectively. We then move on to some common threat models and propose an attack method in sec. \ref{sec: common_attack_models}. Sec. \ref{sec:coma_fedrl} proposes some modifications to the conventional FedRL approach to address the issues of adversaries. Finally in sec. \ref{sec:experimentation} we analyze the proposed attack methods and address them on a series of real-world problems before concluding it in sec. \ref{sec:conclusion}.

\section{Related Work}
\label{sec:related_work}
The effects of adversaries in Machine learning algorithms were first discovered in \cite{szegedy2013intriguing} where it was observed that a small $l_p$ norm perturbation to the input of a trained classifier model resulted in confidently misclassifying the input. These $l_p$ norm perturbations were visually imperceptible to humans. The adversary here acts in the form of specifically creating adversarial inputs to produce erroneous outputs to a learned model \cite{goodfellow2014explaining, kurakin2016adversarial, tramer2017ensemble, kurakin2016adversarial1, moosavi2016deepfool, papernot2016limitations}. For supervised learning problems, such as a classification task, where the network model has already been trained, attacking the input is the most probable choice for an adversary to attack through.
In RL, there is no clear boundary between the training and test phase. The adversary can act either in the form of data-poisoning attacks, such as creating adversarial examples\cite{huang2017adversarial, kos2017delving}, or can directly attack the underlying learned policy \cite{huang2019deceptive, ma2019policy, lin2017tactics, behzadan2018faults} either in terms of malicious falsification of reward signals, or estimating the RL dynamics from a batch data set and poisoning the policy.
Authors in \cite{gleave2019adversarial}, try to attack an RL agent by selecting an adversarial policy acting in a multi-agent environment as a result of creating observations that are adversarial in nature. Their results on a two-player zero-sum game show that an adversarial agent can be trained to interact with the victim winning reliably against it.
In federated RL, alongside the data-poisoning and policy-poisoning attacks, we also have to worry about the model-poisoning attacks. Since we have more than one learning agents, a complete agent can take up the role of an adversary. The adversarial agent can feed in false data to purposely corrupt the global model. In model poisoning attacks the adversary, instead of poisoning the input, tries to adversely modify the learned model parameters directly by feeding false information purposely poisoning the global model \cite{blanchard2017machine, bhagoji2019analyzing}. Since federated learning uses an average operator to merge the local model parameters learned by individual agents, such attacks can severely affect the performance of the global model.

Adversarial training can be used to mitigate the effects of such adversaries. \cite{xie2019feature} showed that the classification model can be made much robust against the adversarial examples by feature de-noising. The robustness of RL policies has also been analyzed by the adversarial training \cite{pinto2017robust, mandlekar2017adversarially, pattanaik2017robust,tessler2019action}.  \cite{pinto2017robust, tessler2019action} show that the data-poisoning can be made a part of RL training to learn more robust policies. They feed perturbed observations during RL training for the trained policy to be more robust to dynamically changing conditions during test time.  \cite{rodriguez2020dynamic} shows that the data-poisoning attacks in federated learning can be resolved by modifying the federated aggregation operator based on induced ordered weighted averaging operators \cite{yager1999induced} and filtering out possible adversaries. 
To the best of our knowledge, there is no detailed research carried out on MT-FedRL in the presence of adversaries. In this paper, we address the effects of model poisoning attacks on the MT-FedRL problem.

\section{Multi-task Federated Reinforcement Learning (MT-FedRL)}
\label{sec:mtrl}

We consider a Multi-task Federated Reinforcement Learning (MT-FedRL) problem with $n$ number of agents. Each agent operates in its own environment which can be characterized by a different Markov decision process (MDP). Each agent only acts and makes observations in its own environment. The goal of MT-FedRL is to learn a unified policy, which is jointly optimal across all of the $n$ environments. Each agent shares its information with a centralized server.  The state and action spaces do not need to be the same in each of these $n$ environments. If the state spaces are disjoint across environments, the joint problem decouples into a set of $n$ independent problems. Communicating information in the case of N-independent problems does not help.

We consider policy gradient methods for RL. The MDP at each agent $i$ can be described by the tuple $\mathcal{M}_i = (\mathcal{S}_i,\mathcal{A}_i,\mathcal{P}_i,\mathcal{R}_i,\gamma_{i})$ 
where $\mathcal{S}_i$ is the state space, $\mathcal{A}_{i}$ is the action space, $\mathcal{P}_{i}$ is the MDP transition probabilities, $\mathcal{R}_{i}:\mathcal{S}_{i}\times\mathcal{A}_{i}\rightarrow \mathbb{R}$ is the reward function, and $\gamma_{i}\in(0,1)$ is the discount factor. 

Let $V^{\pi}_i$ be the value function, induced by the policy $\pi$, at the state $s$ in the $i$-th environment, then we have

\begin{align}
&V_i^{\pi}(s) = \mathbb{E}\left[\sum_{k=0}^{\infty}\gamma_{i}^{k}\mathcal{R}_{i}(s^{k}_{i},a_{i}^{k})\,|\,s_{i}^{0} = s\right],\,\, \notag \\
&a_{i}^{k} \sim \pi(\cdot|s_{i}^{k}).
\label{sec:prob:value_function}  
\end{align}

Similarly, we have the $Q$-function $Q_i^{\pi}$ and advantage function $A_{i}^{\pi}$ for the $i$-th environment as follows

\begin{align}
&Q_i^{\pi}(s_{i},a_{i}) = \mathbb{E}\left[\sum_{k=0}^{\infty}\gamma_{i}^{k}\mathcal{R}(s_{i}^{k},a_{i}^{k})\,|\,s_{i}^{0} = s_{i}, a_{i}^{0} = a_{i}\right],\notag\\ &A_i^{\pi}(s_{i},a_{i}) = Q_i^{\pi}(s_{i},a_{i})- V_i^{\pi}(s_{i}).
\label{sec:prob:QA_function}
\end{align} 

We denote by $\rho_{i}$ the initial state distribution over the action space of $i$-th environment.,
The goal of the MT-FedRL problem is to find a unified policy $\pi^*$ that maximizes the sum of long-term discounted return for all the environment $i$ i.e.

\begin{align}
\max_{\pi} V(\pi;\boldsymbol{\rho}) \triangleq \sum_{i=0}^{n-1}\mathbb{E}_{s_{i}\sim\rho_{i}}V_{i}^{\pi}(s_{i}),
    \quad \boldsymbol{\rho} = \begin{bmatrix}
           \rho_0 \\
           \vdots \\
           \rho_{n-1}
         \end{bmatrix}
\label{sec:prob:obj}
\end{align}
Solving the above equation will yield a unified $\pi^*$ resulting in a balanced performance across all the environments.

We use the parameter $\theta$ to model the family of policies $\pi_\theta(a|s)$, considering both the tabular method (for simpler problems) and neural network-based function approximation (for complex problems). 
The goal of the MT-FedRL problem then is to find $\theta^*$ satisfying
\begin{align}
\theta^* = \arg\max_{\theta} V(\theta;\boldsymbol{\rho}) \triangleq \sum_{i=0}^{n-1}\mathbb{E}_{s_{i}\sim\rho_{i}}V_{i}^{\pi_{\theta}}(s_{i}).   \label{sec:prob:obj_theta}
\end{align}

In tabular method, gradient ascent methods are utilized to solve \eqref{sec:prob:obj} over a set of randomized stationary policies $\{\pi_{\theta}:\theta\in\ \mathbb{R}^{|\mathcal{S}|\times|\mathcal{A}|}\}$, where $\theta$ uses the softmax parameterization
\begin{align}
    \pi_{\theta}(a\,|\,s) = \frac{\exp\left(\theta_{s,a}\right)}{\sum_{a'\in \mathcal{A}}\exp(\theta_{s,a'})}\cdot \label{sec:prob:softmax}    
\end{align}
For a simpler problem where the size of state-space and action-space is limited, this table is easier to maintain.  For more complex problems with a larger state/action space, usually neural network-based function approximation $\{\pi_{\theta}:\mathcal{S} \rightarrow \mathcal{A}\}$ is used, where $\theta$ are the trainable weights of a pre-defined neural network structure.

One approach to solving this unified-policy problem is by sharing the data $\mathcal{M}_i$ observed by each agent in its environment to a centralized server. The centralised server then can train a single policy parameter $\theta$ based on the collective data $\mathcal{M} = \cup_{i=0}^{n-1} \mathcal{M}_i$. This, however, comes with the cost of reduced privacy as the agent needs to share its data with the server.
In MT-FedRL, however, the data tuple $\mathcal{M}_i$ is not shared with the server due to privacy concerns. The data remains at the local agent and instead, the policy parameter $\theta_i$ is shared with the server.
Each agent $i$ utilizes its locally available data $\mathcal{D}_i$ to train the policy parameter $\theta_i$ by maximizing its own local value function $V^{\pi}_i$ through SGD. We assume policy gradient methods for RL training. After the completion of each episode $k$, the agents share their policy parameter $\theta_i^{k-}$ with a centralized server. The server carries out a smoothing average and generates $N$ new sets of parameters $\theta^{k+}_i$, one for each agent, using the following expression. 
\begin{algorithm}[t]
\SetAlgoLined
\textbf{Initialization:} $\theta_i^{0+}\in\mathbb{R}^d$, step sizes $\delta^k$, smoothing average threshold iteration $t$

 \texttt{\%Server Executes} \newline
  \For{k=1,2,3,...}{
   Calculate smoothing average parameters
    \begin{align*}
     \alpha^k = \frac{1}{n} \max(1, k/t), \quad
     \beta^k = \frac{1-\alpha^k}{n-1} 
    \end{align*}

    \For{each agent i \textbf{in parallel}}{
          Receive updated policy parameter from clients
          $\theta^{(k+1)-}_{i} \leftarrow \texttt{ClientUpdate}\left(i,\theta^{k+}_{i}\right)$
          }
    \For{each agent i}{
        Policy update:
            \begin{align*}
            \theta^{(k+1)+}_{i} &= \alpha^k \theta^{(k+1)-}_{i} + \beta^k \sum_{i\neq j}\theta^{(k+1)-}_{j}
            \end{align*}
        Send updated policy parameter $\theta^{(k+1)+}_{i}$ back to client $i$  
    }
    }
    \SetKwFunction{FMain}{ClientUpdate}
    \SetKwProg{Pn}{Function}{:}{}
    \Pn{\FMain{$i$, $\theta$}}{
    1) Compute the gradient of the local value function $\frac{\partial V_{i}^{\pi_{\theta}}(\rho_i)}{\partial \theta_{s_{i},a_{i}}}$ based on the local data
    
    2) Update the policy parameter
    \begin{align}
        \theta^- = \theta + \delta^k \frac{\partial V_{i}^{\pi_{\theta}}(\rho_i)}{\partial \theta_{s_{i},a_{i}}}
    \end{align}
    }
    
    \KwRet  $\theta^{-}$
    
\caption{Multi-task Federated Reinforcement Learning (MT-FedRL) with smoothing average}
\label{Alg:MTRL}
\end{algorithm}


\begin{equation}
    \theta^{k+}_i = \alpha^k \theta_i^{k-} + \beta^k \sum_{j \neq i} \theta_j^{k-}
    \label{eq:smooth_avg}
\end{equation}
where $\alpha^k, \beta^k=\frac{1-\alpha}{n-1} \in (0, 1)$ are non-negative smoothing average weights. The goal of this smoothing average is to achieve a consensus among the agents' parameters, i.e.

\begin{equation}
    \lim_{k\rightarrow \infty} \theta^{k+}_i \rightarrow \theta^*  \quad \forall i \in \{0, n-1\}
    \label{eq:convergence}
\end{equation}

As the training proceeds, the smoothing average constants converge to $\alpha^k, \beta^k \rightarrow \frac{1}{n}$. The conditions on $\alpha^k, \beta^k$ to guarantee the convergence of Algorithm \ref{Alg:MTRL} can be found in \cite{zeng2020decentralized}. The complete algorithm of multi-task federated RL can be found in Alg. \ref{Alg:MTRL}

\section{MT-FedRL with adversaries}
\label{sec:mtrl_adv}
MT-FedRL has proven to converge to a unified policy that performs jointly optimal on each environment \cite{zeng2020decentralized}. This jointly optimal policy yields near-optimal policies when evaluated on each environment if the agents' goals are positively correlated. If the agent's goals are not positively correlated, the unified policy might not result in a near-optimal policy for individual environments. This is exactly what happens to MT-FedRL in the presence of an adversary.

Let $\mathcal{L}$ denote the set of adversarial agents in a $n-agent ~MTFedRL$ problem. 
The smoothing average at the server can be decomposed based on the adversarial and non-adversarial agent as follows
\begin{equation}
    \theta^{k+}_i = \alpha^k \theta_i^{k-} + \beta^k \sum_{j \neq i, j \notin \mathcal{L}} \theta_j^{k-} + \beta^k \sum_{l \in \mathcal{L}} \theta_l^{k-}
    \label{eq:smoothing_avg_adv}
\end{equation}

where $i\notin \mathcal{L}$. $\theta^{k+}_i$ is the updated policy parameter for agent $i$ calculated by the server at iteration $k$. This update incorporates the knowledge from other environments and as the training proceeds, these updated policy parameters for all the agents converge to a unified parameter $\theta^*$. In a non-adversarial MT-FedRL problem, this unified policy parameter ends up achieving a policy that maximizes the sum of discounted returns for all the environments. In an adversarial MT-FedRL problem, the goal of the adversarial agent is to prevent the MT-FedRL from achieving this unified $\theta^*$ by purposely providing an adversarial policy parameter $\theta_l^{k-}$. 

\bigbreak
\noindent \textbf{Parameters that effect learning:} \label{sec:parameters}
Using gradient ascent, each agent updates its own set of policy parameter locally according to the following equation,
\begin{align}
    \theta_i^{k-} = \theta_i^{(k-1)+} + \delta_i \nabla_{\theta_i}V_{i}^{\pi_{\theta_i}}(\rho_{i})
    \label{eq:local_update}
\end{align}

where $\delta_i$ is the learning rate for agent $i$. Using Eq. \ref{eq:local_update} in the smoothing average Eq. \ref{eq:smooth_avg} yields
\begin{align}
    \theta^{k+}_i = \left( \alpha^k \theta_i^{(k-1)+} + \beta^k \sum_{j \neq i, j \notin \mathcal{L}} \theta_j^{(k-1)+} \right) + \notag \\
     \left( \alpha^k \delta_i  \nabla_{\theta_i}V_{i}^{\pi_{\theta_i}}(\rho_{i}) + \beta^k \sum_{j \neq i, j \notin \mathcal{L}}  \delta_j \nabla_{\theta_j}V_{j}^{\pi_{\theta_j}}(\rho_{j}) \right) + \notag \\
    \left( \beta^k \sum_{l \in \mathcal{L}} \theta_l^{k-} \right)
    \label{eq:compromise}
\end{align}

The server update of the policy parameter can be decomposed into three parts. 
\begin{itemize}
    \item The weighted sum of the previous set of policy parameters $\theta_i^{(k-1)+}$ shared by the server with the respective agents. 
    \item The agent's local update, which tries to shift the policy parameter distribution towards the goal direction.
    \item The adversarial policy parameter which aims at shifting the unified policy parameter away from achieving the goal.
\end{itemize}

If the update carried out by the adversarial agent is larger than the sum of each agent's policy gradient update, the policy parameter will start moving away from the desired consensus $\theta^*$. The success of the adversarial attack hence depends on,

\begin{itemize}
    \item The nature of adversarial policy parameter $\theta_l^{k-}$
    \item Non-adversarial agent's local learning rate $\delta_i$
    \item The number of non-adversarial agents $n-|\mathcal{L}|$
\end{itemize}
An adversarial attack is more likely to be successful if the local learning rate of non-adversarial agents $\delta_i$ is small and the number of adversarial agents $|\mathcal{L}|$ is large.

\bigbreak
\noindent \textbf{Threat Model:}
For an adversarial agent to be successful in its attack, it needs to shift the convergence from $\theta^*$ in Eq. \ref{eq:convergence} to $\theta^\prime$ such that the resultant policy $\pi^\prime$ follows
\begin{align}
\mathbb{E}_{s_{i}\sim\rho_{i}}V_{i}^{\pi^\prime}(s_{i}) << \mathbb{E}_{s_{i}\sim\rho_{i}}V_{i}^{\pi^*}(s_{i}), \quad \forall i\notin \mathcal{L}
\label{eq:adv_goal}
\end{align}

\begin{figure}
  \includegraphics[width=\linewidth]{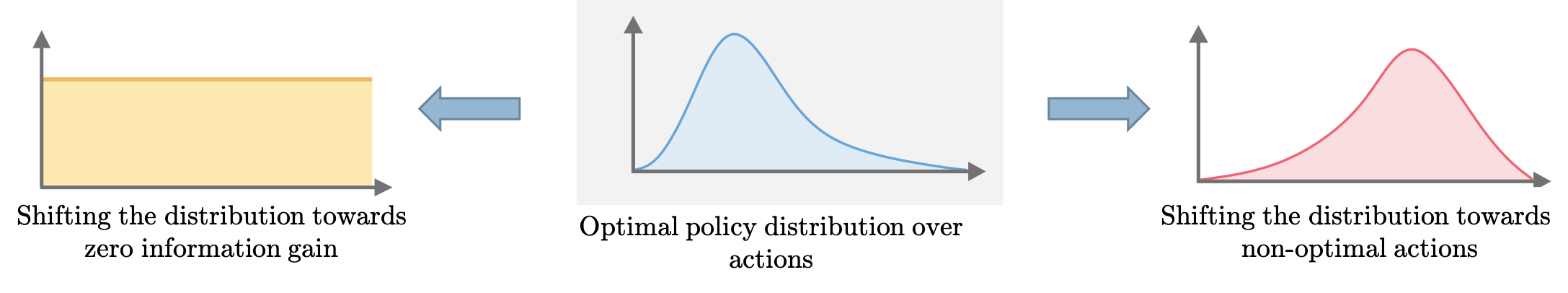}
  \caption{The objective of an adversarial agent is to shift the policy distribution that yields poor actions}
  \label{fig:shift_dist}
\end{figure}

The only way an adversarial agent can control this convergence is through the policy parameter $\theta_l^{k-}$ that it shares with the server. The adversarial agent needs to share the policy parameter that moves the distribution of the smoothing average of non-adversarial agents either to a uniform distribution or in the direction that purposely yields bad actions (Fig. \ref{fig:shift_dist}). Generally shifting the distribution to uniform distribution will require less energy than to shift it to a non-optimal action distribution. This requires that the adversary cancel out the information gained by all the other non-adversarial agents hence not being able to differentiate between good and bad actions, leaving all the actions equally likely. 

\begin{figure}
  \includegraphics[width=\linewidth]{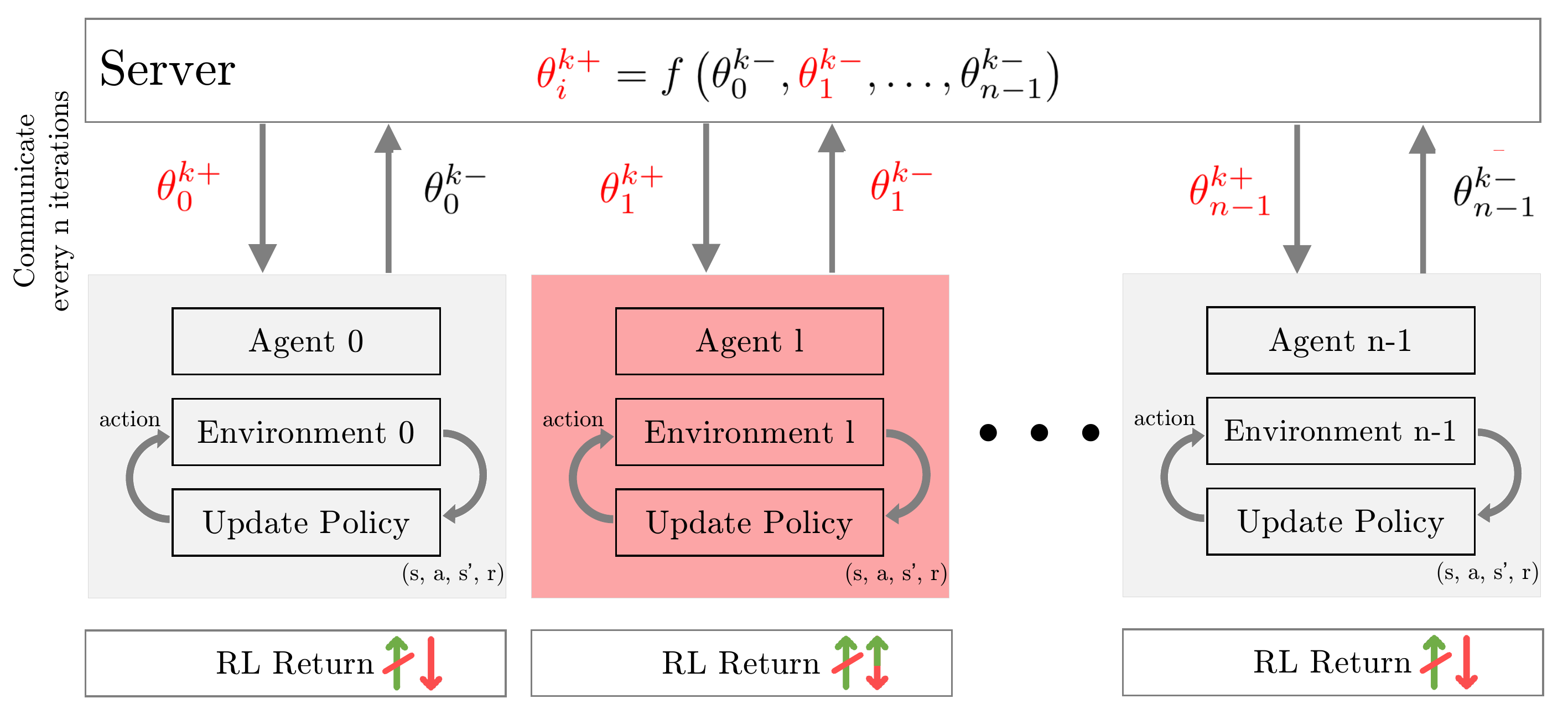}
  \caption{Adversaries can negatively impact the unified policy by providing adversarial policies to the server. This results in negatively impacting the achieved discounted return on the environments}
  \label{fig:block_adv}
\end{figure}

\bigbreak
\noindent \textbf{Threat Model:} \label{sec:threat_model}
We will assume the following threat model. At iteration $k$, each adversarial agent $l$ shares the following policy parameter with the server
\begin{align}
    \theta_l^{k-} = \lambda^k \theta_{adv}^k
\end{align}
Hence the threat model is defined by the choice of the attack model $\theta_{adv}$ and $\lambda^k \in \mathbb{R}$ which is a non-negative iteration-dependant scaling factor that will be used to control the norm of the adversarial attack. To make the scaling factor more meaningful, we will assume that
\begin{align*}
    \|\theta_{adv}\|^2 \approx \frac{1}{\left(n-|\mathcal{L}|\right)}\sum_{i\notin\mathcal{L}}\|\theta_{i}\|^2
\end{align*}
The relative difference in the norm of the policy parameter between the adversarial agent and non-adversarial agent will be captured in the scaling factor term $\lambda^k$. One thing we need to be mindful of is the value of the scaling factor. If the scaling factor is large enough, almost any random (non-zero) values for the adversarial policy parameter $\theta_l^k$ will result in a successful attack. 
We will quantify the relative performance of the threat models by 
\begin{itemize}
    \item How effective they are in attacking, either in terms of the achieved discounted return under the threat induced unified policy or in terms of a more problem-specific goal parameter (more in the experimentation section).
    \item If two threat models achieve the same attacking performance, the one with a smaller scaling factor $\lambda^k$ will be considered better. The smaller the norm of the adversarial policy parameter, the better the chances for the threat model to go unnoticed by the server.
\end{itemize}

\section{Common Attack Models}
\label{sec: common_attack_models}

In this section, we will discuss a few common attack models $\theta_{adv}$ and propose an adaptive attack model. For the rest of the section, we will focus on the single-agent adversarial model $|\mathcal{L}|=1$. The extension of these threat models to multiple adversarial agents is straight forward.
\subsection{Random Policy Attack (Rand)}
This attack will be used as a baseline for the other attack methods. In a Random policy attack, the adversarial agent maintains a set of random policy parameter sampled from a Gaussian distribution with mean 0 and standard deviation $\sigma \in \mathbb{R}$ i.e. for each element $\theta_{adv,j}$ of $\theta_{adv}$
\begin{align}
    \theta_{adv,j} &\sim \mathcal{N}(0, \sigma)
\end{align}
This attack assumes that the adversary has no knowledge to estimate the best attack method from. If the scaling factor $\lambda^k$ is large enough, this attack method can shift the distribution of the policy towards a random distribution.
\subsection{Opposite Goal Policy Attack (OppositeGoal)}
This attack method assumes that a sample environment is available for the agent to devise the attack. In this attack method, the adversary $l$ learns a policy $\pi^{OG}_{\theta_{adv}}$ utilizing its local environment with the goal of minimizing (instead of maximizing) the long term discounted return i.e. the objective function to maximize is
\begin{equation}
    J(\theta_{adv}) = -V_{l}^{\pi_{\theta_{adv}}}(\rho_{l})
    \label{eq:OG}
\end{equation}

With the completion of an episode $k$, the adversary updates its policy parameter $\theta_{adv}$ locally by maximizing Eq.\ref{eq:OG} and shares the scaled version of the updated policy parameter with the server.

The OppositeGoal attack method can either shift the policy to a uniform distribution, or to a distribution which prefers actions that yield opposite goal. For the agent to shift the distribution to uniform, the following constraints need to hold.

\begin{enumerate}
    \item All the N environments should be similar enough that they generate policies that are close enough i.e.
    \begin{equation}
    \frac{1}{|\mathcal{S}|}\sum_{s\in\mathcal{S}} KL(\pi_{\theta_i}(.|s), \pi_{\theta_j}(.|s)) \leq \epsilon
    \end{equation}
    or equivalently if the learning rate and initialization is the same for each agent then

    \begin{equation}
        \left\| \frac{\theta_i}{\|\theta_i\|_2} - \frac{\theta_j}{\|\theta_j\|_2}\right\|_2 \leq \epsilon
    \end{equation}
    
    \item Action selection based on minimum probability action for OppositeGoal policy should be close enough to action selection based on maximum probability for NormalGoal Policy 
    \begin{equation}
    \frac{1}{|\mathcal{S}|}\sum_{s\in\mathcal{S}} KL(1-\pi_{\theta_i}^{OG}(.|s), \pi_{\theta_j}(.|s)) \leq \epsilon
    \end{equation}
    or equivalently if the learning rate is same initialization is zero $\theta_i^0 = \mathbf{0}$

        \begin{equation}
        \left\| \frac{\theta_i}{\|\theta_i\|_2} + \frac{\theta_j}{\|\theta_j\|_2}\right\|_2 \leq \epsilon
    \end{equation}
    
\end{enumerate}
In short, all the $N$ environments should be similar enough such that training on an opposite goal will yield a policy that when combined with a policy learned to maximize the actual goal will yield a complete information loss. 

Most of the time, these assumptions will not hold as they are too strict (the difference in environment dynamics, initialization of policy parameter, the existence of multiple local minima, etc.). Instead, if the scaling factor is large, the OppositeGoal attack will shift the distribution of the consensus to an opposite goal policy. Since we are taking into account the environment dynamics, this attack will however be better than the random policy attack.

\subsection{Adversarial Attack by Minimizing Information Gain (\textit{AdAMInG})}

Even though the adversarial choice of opposite goal makes an intuitive sense as the best attack method, we will see in the results section that it's not.
Hence, we propose an attack method that takes into account the nature of MT-FedRL smoothing averaging and devises the best attack given the information available locally. The goal of \textit{AdAMInG} is to devise an attack that uses a single adversarial agent with a small scaling factor by forcing the server to forget what it learns from the non-adversarial agents.

For the smoothing average at the server to lose all the information gained by other non-adversarial agents we should have
\begin{align}
    \theta_l^{k-} = -\frac{1}{\beta^k |\mathcal{L}|} \left( \alpha^k \theta_i^{k-} + \beta^k \sum_{j \neq i, l} \theta_j^{k-} \right)
\end{align}
Using the above equation in Eq. \ref{eq:smoothing_avg_adv} will result $\theta_i^{k+} = \mathbf{0}$, hence losing the information gained by $\theta_i^{k-}$. The problem in the above equation is that the adversarial agents do not have access to the policy parameter shared by non-adversarial agents $\theta_i^{k-},  \forall i \neq l$ and hence the quantity in the parenthesis (smoothing average of the non-adversarial agents) is unknown. The attack model then is to estimate the smoothing average of the non-adversarial agents.

The adversarial agent has the following information available to it
\begin{itemize}
    \item The last set of policy parameter shared by the adversarial agent to the server $\theta_l^{(k-1)-}$
    \item The federated policy parameter shared by the server to the adversarial agent $\theta_l^{(k-1)+}$
\end{itemize}

The adversarial agent can estimate the smoothing average of the non-adversarial agents from these quantities. The \textit{AdAMInG} attack shares the following policy parameter

\begin{align}
    \theta_l^{k-} = \lambda^k \left( \frac{ \alpha^k \theta_l^{(k-1)+} - \theta_l^{(k-1)-}}{\beta^k }\right)
    \label{eq:adaming_adv}
\end{align}

The smoothing average at the server for $i\in\{0, n-1\}, i\neq l$ becomes
\begin{align*}
    \theta_i^{k+} &= \alpha^k\theta^{k-}_i + \beta^k \sum_{i\neq j,l}\theta_j^{k-}+\beta^k\theta_l^{k-} \notag \\
    &= \alpha^k\theta_i^{k-} +\beta^k \sum_{i\neq j,l}\theta_j^{k-}- \\
    & \quad \frac{\lambda^k}{n-1} \alpha^k \left(\theta_l^{(k-1)+} - \theta_l^{(k-1)-}\right) \notag \\
    &= \alpha^k\theta_i^{k-} +\beta^k \sum_{i\neq j,l}\theta_j^{k-}-\frac{\lambda^k}{n-1}\beta^k \sum_{j\neq l} \theta_j^{(k-1)-} \notag \\
    &= \left(\alpha^k\theta_i^{k-} - \frac{\lambda^k}{n-1}\beta^k\theta^{(k-1)-}_i\right) \notag \\&\quad \quad \quad \quad + \sum_{j\neq i, l}\left( \beta^k\theta_j^{k-} - \frac{\beta^k\lambda^k}{n-1} \theta_j^{(k-1)-}\right)
\end{align*}
We want $\theta_i^{k+} \rightarrow \mathbf{0}, ~\forall i\in\{0, n-1\}, i\neq l$. This means forcing the two terms inside the parenthesis to $\mathbf{0}$. If the initialization of all the agents are same, i.e. $\theta_i^{0-}=\theta^{0}=\mathbf{0} ~\forall i$ and the learning rate is small, we have $\|\theta_i^{k-} - \theta_i^{(k-1)-}\| < \epsilon$. Hence $\theta_i^{k+} \rightarrow \mathbf{0} $ can be achieved by the following scaling factor
\begin{align*}
    \lambda^{k*} = argmin_{\lambda^k} ~ g(\lambda^k, n) \vspace{-1em}
\end{align*}

where
\begin{align*}
    g(\lambda^k, n) = ~ \left|\alpha^k - \beta^k \frac{\lambda^k}{n-1}\right| + \left|\beta^k \left(1-\frac{\lambda^k}{n-1}\right)(n-2)\right|
\end{align*}
For simplicity we have not shown the dependence of $\alpha^k, \beta^k$ in the expression $g(\lambda^k, n)$ as they directly depend on $k$. Solving this optimization problem yields
\begin{align}
    \lambda^* = n-1, \quad  n\geq3
    \label{eq:lambda}
\end{align}

This means that the scaling factor should be equal to the number of non-adversarial agents and is independent of the iteration $k$. For $\lambda^k < \lambda^*$ we still can achieve a successful attack if the learning rate $\delta$ is not too high. 

As the training proceeds, the values of the smoothing constants $\alpha^k, \beta^k$ approach their steady-state value of $\frac{1}{n}$. At that point, the steady-state value of $g(\lambda^k, n)$ defined as $g_{ss}(\lambda^k, n)$ is given by
\begin{align}
    g_{ss}(\lambda^{k}, n) = \frac{n-1-\lambda}{n}
    \label{eq:g_ss}
\end{align}

\begin{figure}[t]
  \includegraphics[width=\linewidth]{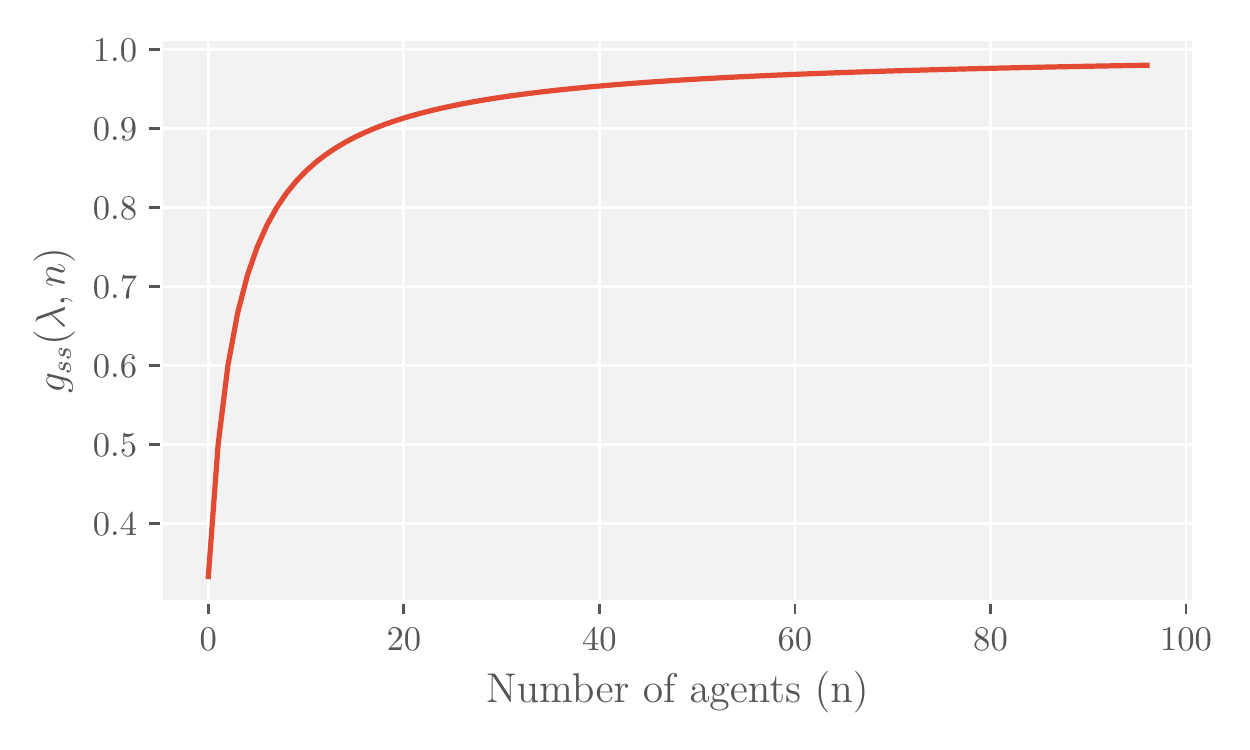}
  \caption{$g\left(\lambda^k, n\right)$ as a function of $\lambda^k=1$ and $n$}
  \label{fig:gk_graph_n}
\end{figure}

\begin{figure}[t]
  \includegraphics[width=\linewidth]{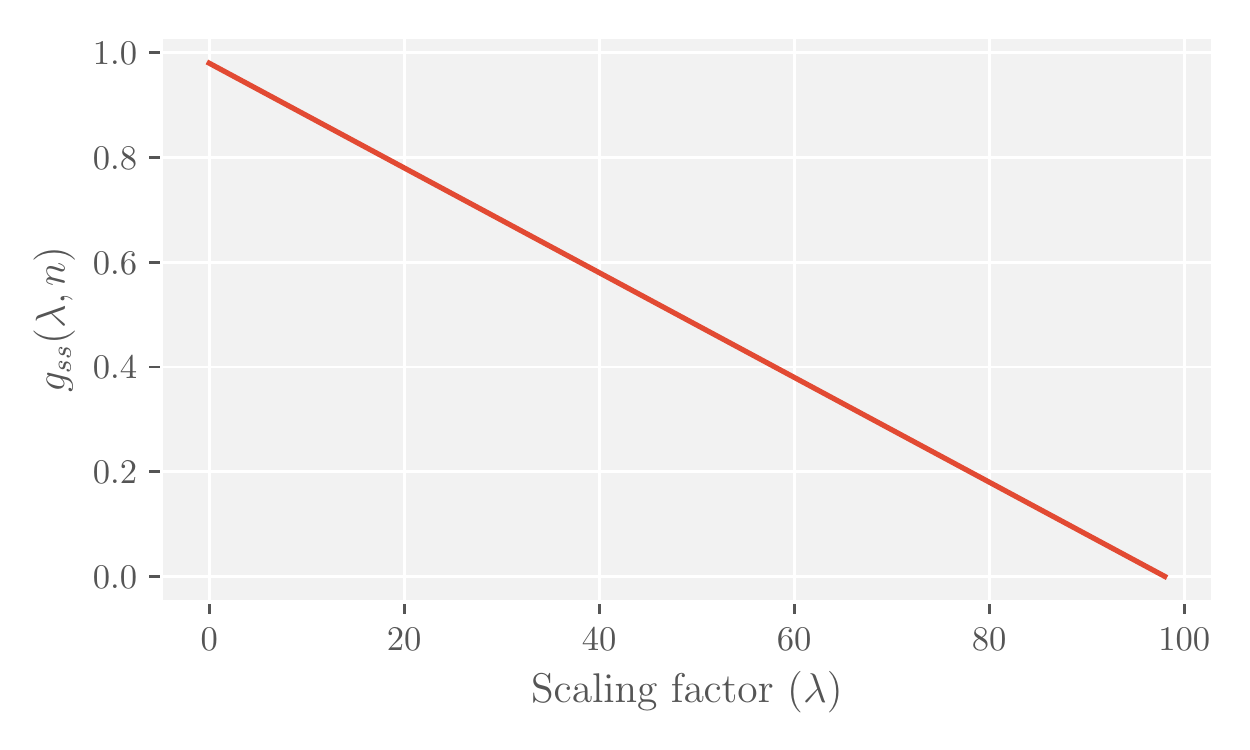}
  \caption{$g\left(\lambda^k, n\right)$ as a function of $\lambda^k$ and $n=100$}
  \label{fig:gk_graph_lambda}
\end{figure}

The steady-state value $g_{ss}(\lambda^{k}, n)$ signifies how effective/successful the \textit{AdAMInG} attack will be for the selected parameters ($\lambda^k, n$). A steady-state value of 0 signifies a perfect attack, where the policy parameter shared by the server loses all the information gained by the non-adversarial agents. On the other hand, a steady-state value of 1 indicates a completely unsuccessful attack. The smaller the $g_{ss}(\lambda^{k}, n)$, the better the \textit{AdAMInG} attack.

Fig. \ref{fig:gk_graph_n} plots $g(\lambda^k, n)$ as a function of the number of agents $n$ for a scaling factor of 1 ($\lambda^k=1$). It can be seen that as the number of agents increases, the steady-state value $g_{ss}$ becomes closer to 1 making it difficult for \textit{AdAMInG} to have a successful attack with a scaling factor of 1. As the number of agents increases, the update carried out by the non-adversarial agent has a more significant impact on the smoothing average than the adversarial agent making it harder for the adversarial agent to attack. 
Fig. \ref{fig:gk_graph_lambda} plots $g(\lambda^k, n)$ as a function of the scaling factor $\lambda^k$ for $n=100$. It can be seen that the scaling factor has a linear impact on the success of the \textit{AdAMInG} attack. The performance of the \textit{AdAMInG} attack increases linearly with the increase in the scaling factor. The best \textit{AdAMInG} attack is achieved when $\lambda^k = n-1$ which is consistent with Eq. \ref{eq:lambda}.

In the experimentation section, we will see that a non-zero steady-state value can still result in a successful attack for a small learning rate $\delta$.

It is safe to assume that if we do not change the learning rate (and it is small enough), we can find the scaling factor required to achieve the same attacking performance by increasing the number of agents $n$. The steady-state relationship between $n$ and $\lambda$ in Eq. \ref{eq:g_ss} lets us analyze the relative attacking performances by varying the number of agents $n$. Let's say that for $n_1$ number of agents and a given learning rate that is small, we were able to achieve a successful attack with $\lambda_1$. Now to achieve the same successful attack for $n_2$ number of agents we need
\begin{align}
    \lambda_2 = \frac{n_2(1+\lambda_1)}{n_1}-1
    \label{eq:lambda_n_relation}
\end{align}

Unlike the OppositeGoal attack, we can guarantee that the \textit{AdAMInG} attack will yield a successful attack if the scaling factor is equal to the number of non-adversarial agents.

We will see in the results section that the scaling factor needs no to be this high if the learning rate $\delta$ is not high. We will be able to achieve a good enough attack even if $\lambda^k<n-1$.
The only down-side with the \textit{AdAMInG} attack method is that it requires twice the amount of memory as compared to that of OppositeGoal or Rand attack method. \textit{AdAMInG} attack method needs to store the both the adversary shared policy parameter $\theta_l^{(k-1)-}$ and the server shared policy parameter $\theta_l^{(k-1)+}$ from the previous iteration to compute the new set of policy parameters to be shared $\theta_l^{k-}$ as shown in Eq. \ref{eq:adaming_adv}. However, as opposed to the OppositeGoal attack method, the \textit{AdAMInG} attack method does not require to learn from the data sampled from environment saving up much on the compute cost.

\section{Detecting attacks - ComA-FedRL}
\label{sec:coma_fedrl}
\begin{table}[t]
\centering
\begin{tabular}{ccc}
\toprule
\textbf{Evaluated Policy} &  \textbf{Environment} &  \textbf{Cumulative reward} \\ \bottomrule
Non-adv                   &  Non-adv              &  High                     \\
Non-adv                   &  Adv                  &  Low (Secondary Attack)     \\
Adv                       &  Non-adv              &  Low                        \\
Adv                       &  Adv                  &  Low (Secondary Attack)    \\\bottomrule
\end{tabular}
\caption{Cross-evaluation of policies in \texttt{ComA-FedRL} in terms of cumulative return}
\label{tab:cross_eval}
\end{table}

\begin{algorithm}
\SetAlgoLined
\textbf{Initialization:} {\small Initialize number agents $n$, $\theta_i^0\in\mathbb{R}^d$, step size $\delta^k$, 
$base\_comm\in\mathbb{R},~comm[i] = base\_comm ~\forall i \in \{0, n-1\},~wait\_comm\in\mathbb{R}$}

  \For{k=1,2,3,...}{
  \texttt{\% Pre-train phase} \newline
  \If{$k \leq wait\_comm$}
  {
    \If{$k\% base\_comm=0$}
    {
        \For{each agent i \textbf{in parallel}}{
          $\theta^{(k+1)-}_{i} \leftarrow \texttt{ClientUpdate} \left(i,\theta^{k+}_{i}\right)$
          }
        $r \leftarrow \texttt{CrossEvalPolicies}\left(r, \theta^{(k+1)-}\right)$
    }
  }
  \Else{
   
  Calculate smoothing average parameters $\alpha_k, \beta_k$
$comm \leftarrow$ \texttt{UpdateCommInt}($r, comm$)
    
    \For{each agent i \textbf{in parallel}}{
        \If{$k\%comm[i]=0$}{
          $\theta^{(k+1)-}_{i} \leftarrow$ \texttt{ClientUpdate}$\left(i,\theta^{k+}_{i}\right)$
          }
          }

    $num\_active\_agents=0$  \newline        
    \For{each agent i}{
    \If{$k\%comm[i]=0$}{
            num\_active\_agents+=1
            \begin{align*}
            \theta^{(k+1)+}_{i} &= \alpha^k \theta^{(k+1)-}_{i} + \beta^k \sum_{i\neq j}\theta^{(k+1)-}_{j}
            \label{Alg:Distributed_PG:Update}
            \end{align*}
        Send $\theta^{(k+1)+}_{i}$ back to client $i$  
    }
    }
    \If{$num\_active\_agents=n$}
    {
    $r \leftarrow \texttt{CrossEvalPolicies}\left(r, \theta^{(k+1)-}\right)$
    }
    }
}
 ~ \newline
\SetKwFunction{FMain}{CrossEvalPolicies}
\SetKwProg{Pn}{Function}{:}{}
\Pn{\FMain{r, $\theta$}}{
        \For{each agent i \textbf{in parallel}}{
        Randomly assign each agent $i$ another agent $j$ without replacement \\
          $r_{j}.append(\texttt{ClientEvaluate}(i,\theta_{j}))$
          }
         }
\KwRet  $r$

\SetKwFunction{FMain}{ClientUpdate}
\SetKwProg{Pn}{Function}{:}{}
\Pn{\FMain{$i$,$\theta$}}{
1) Compute the gradient of the local value function $\frac{\partial V_{i}^{\pi_{\theta}}(\rho_i)}{\partial \theta_{s_{i},a_{i}}}$ based on the local data;

2) Update the policy parameter
\begin{align*}
    \theta^{-} = \theta + \delta^k \frac{\partial V_{i}^{\pi_{\theta}}(\rho_i)}{\partial \theta_{s_{i},a_{i}}}
\end{align*}
}
    
\KwRet  $\theta^{-}$

\SetKwFunction{FMain}{ClientEvaluate}
\SetKwProg{Pn}{Function}{:}{}
\Pn{\FMain{$i$, $\theta_{j}$}}{
Evaluate the policy $\theta_j$ on agent $i$ and return the cumulative reward $ret$
}
\KwRet  $ret$

\caption{Communication Aware Federated RL}
\label{Alg:comafedRL}
\end{algorithm}

\begin{figure*}[t]
\centering
  \includegraphics[width=0.9\linewidth]{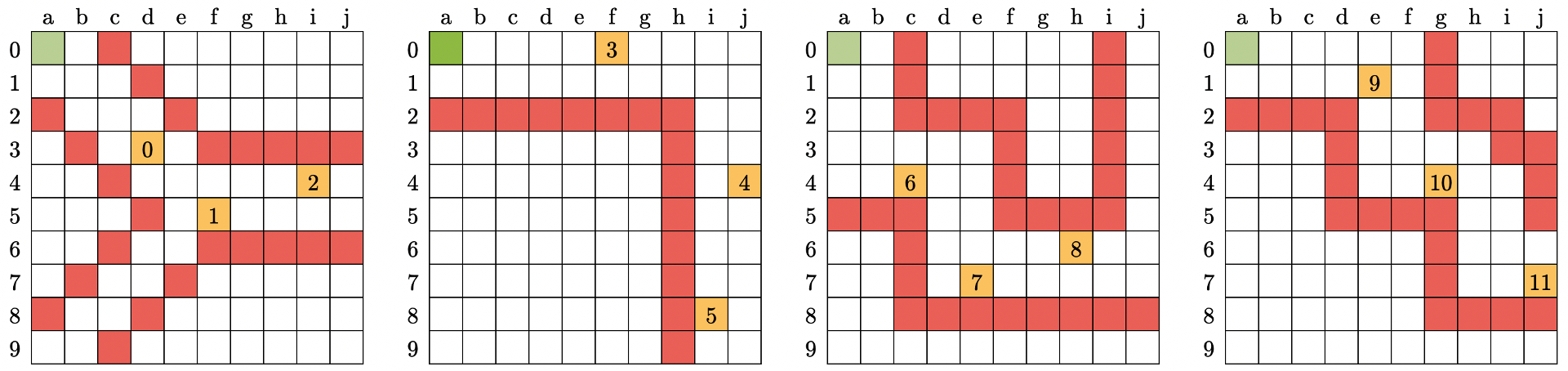}
  \caption{[GridWorld] The 12 environments used}
  \label{fig:gridworld_maze}
\end{figure*}

We will see in Sec. \ref{sec:experimentation} that the FedRL algorithm under the presence of an adversary can severely affect the performance of the unified policy. Hence, we propose Communication Adaptive Federated RL (\texttt{ComA-FedRL}) to address the adversarial attacks on a Federated RL algorithm. Instead of communicating the policy parameter from all agents at a fixed communication interval, we assign different communication intervals to agents based on the confidence of them being an adversary. An agent, with higher confidence of being an adversary, is assigned a large communication interval and vice-versa. Communicating less frequently with an adversary agent can greatly mitigate its effects on the learned unified policy. Since we can't guarantee that a certain agent is an adversary or not, we can't just cut off the communication with an agent we think would be an adversary. Moreover, an adversary can fake being a non-adversarial agent to get away with being marked as an adversarial agent. Hence, we don't mark agents as adversary or non-adversary, rather we adaptively vary the communication interval between the server and the agents based on how good, on average, does the policy of the agent performs in other environments. The complete algorithm of \texttt{ComA-FedRL} can be seen in Algo. \ref{Alg:comafedRL}.

\texttt{ComA-FedRL} begins with a pre-train phase, where each agent tries to learn a locally optimistic policy independent of others. These locally optimistic policies are expected to perform better than a random policy on other agents' environments. After every certain number of episodes, the server randomly assigns a policy to all the environments without replacement for evaluation, and the cumulative reward achieved by this policy is recorded. Based on the nature of the policy and the environment it is cross-evaluated in, we have four cases as shown in Table \ref{tab:cross_eval}. When the policy locally learned by a non-adversarial agent is evaluated in the environment of a non-adversarial agent, it generally performs better than a random policy because of the correlation of the underlying tasks. Hence we get a slightly higher cumulative reward compared to other cases. On the other hand, if an adversarial policy is cross-evaluated on a non-adversarial agent's environment, it generally performs worse because of the inherent nature of the adversary, giving a low cumulative reward. When the policies are evaluated on the adversarial agent's environment, the adversary can present a secondary attack in faking the cumulative reward. It intentionally reports a low cumulative return with the hopes of confusing the server to mistake a non-adversarial agent with an adversarial one. Since the adversarial agent has no way of knowing if the policy shared by the server belongs to an adversarial or a non-adversarial agent, it always shares a low cumulative return.

At the end of the pre-train phase, the cumulative rewards are averaged out for a given policy and are compared to a threshold. If the averaged cumulative reward of the policy is below (above) this threshold, the policy is marked as \textit{possibly-adversarial} (\textit{possibly-non-adversarial}). The \textit{possibly-adversarial} agents are assigned a higher communication interval (less frequent communication), while \textit{possibly-non-adversarial} agents are assigned a smaller communication interval (more frequent communication). The agents are constantly re-evaluated after a certain number of iterations and the categories associated with the agents are updated. After re-evaluation, if an already marked possible-adversary agent is re-marked as \textit{possibly-adversary}, the agent's communication interval is doubled, signifying a higher probability of it being an adversary and making it contribute even lesser towards the server smoothing average. Hence as the training proceeds, the adversarial agent's contribution to the server smoothing average becomes smaller and smaller. 

Further details on the variables and functions used in Alg. \ref{Alg:comafedRL} can be found in the Appendix section \ref{sec:training_details}.

\section{Experimentation}
\label{sec:experimentation}
 For the entire experimentation section, we focus on single-adversary MT-FedRL and hence $|\mathcal{L}|=1$. We report the experimental results from a simpler tabular-based RL problem (GridWorld) to a more complex neural network-based RL problem (AutoNav). In both cases, we use policy gradient based RL methods.

\subsection{GridWorld - Tabular RL}

\noindent \textbf{Problem Description:}  We begin our experimentation with a simple problem of GridWorld. The environments are grid world mazes of size $10 \times 10$ as seen in Fig. \ref{fig:gridworld_maze}. Each cell in the maze is characterized into one of the following 4 categories: \textit{hell-cell} (red), \textit{goal-cell} (yellow), \textit{source-cell} (green), and \textit{free-cell} (white). The agent is initialized at the \textit{source-cell} and is required to reach the \textit{goal-cell} avoiding getting into the \textit{hell-cell}. The \textit{free-cells} in the maze can be occupied without any consequences.  The agent can take one of the following 4 actions $ \mathcal{A} = \{\texttt{move-up}, ~\texttt{move-down}, ~\texttt{move-right},~ \texttt{move-left}\}$ which corresponds to the agent moving one cell in the respective direction. At each iteration, the agent observes a one-step SONAR-based state $s\in\mathcal{R}^4$ which corresponds to the nature of the four cells (up, down, right, left) surrounding the agent. If the corresponding cell is a \textit{hell-cell}, \textit{goal-cell}, or \textit{free-cell}, the corresponding state element is -1, 1, or 0 respectively. Hence, we have $|\mathcal{S}| = 81$. Based on the nature of the environment, only a subset of these states will be available for each environment. 
At each iteration, the agent samples an action from the action space and based on the next state, observes a reward. The reward is -1, 1, 0.1, or -0.1 if the agent crashed into hell-cell, reached the goal, moved closer to or away from the goal respectively. The effectiveness of the MT-FedRL-achieved unified policy is quantified by the win ratio ($WR$) defined by
\begin{align*}
    WR = \frac{1}{n-1}\sum_{i\neq l} \frac{\text{\# of times agent $i$ reached goal state}}{\text{total \# of attempts in environment $i$}}
\end{align*}
In this 12-agent MT-FedRL system, agent 0 is assigned the adversarial role ($l=0$). The goal for agent 0 is to decrease this win ratio. We will characterize the performance of the adversarial attack by the probability of successful attack $p_{sa}$ given by
\begin{align*}
    p_{sa} = 1 -  \frac{WR_{adv}}{WR_{no-adv}}
\end{align*}
where $WR_{adv}$ is the win ratio with an adversary, while $WR_{no-adv}$ is the win ratio without any adversary. An attack method is said to be successful with probability $p_{sa}$ if it is able to defeat the system $p_{sa}\%$ of the time compared to a non-adversarial MT-FedRL. The greater the $p_{sa}$ the better the attack performance of the adversary.

\begin{figure}[t]
\centering
  \includegraphics[width=\linewidth]{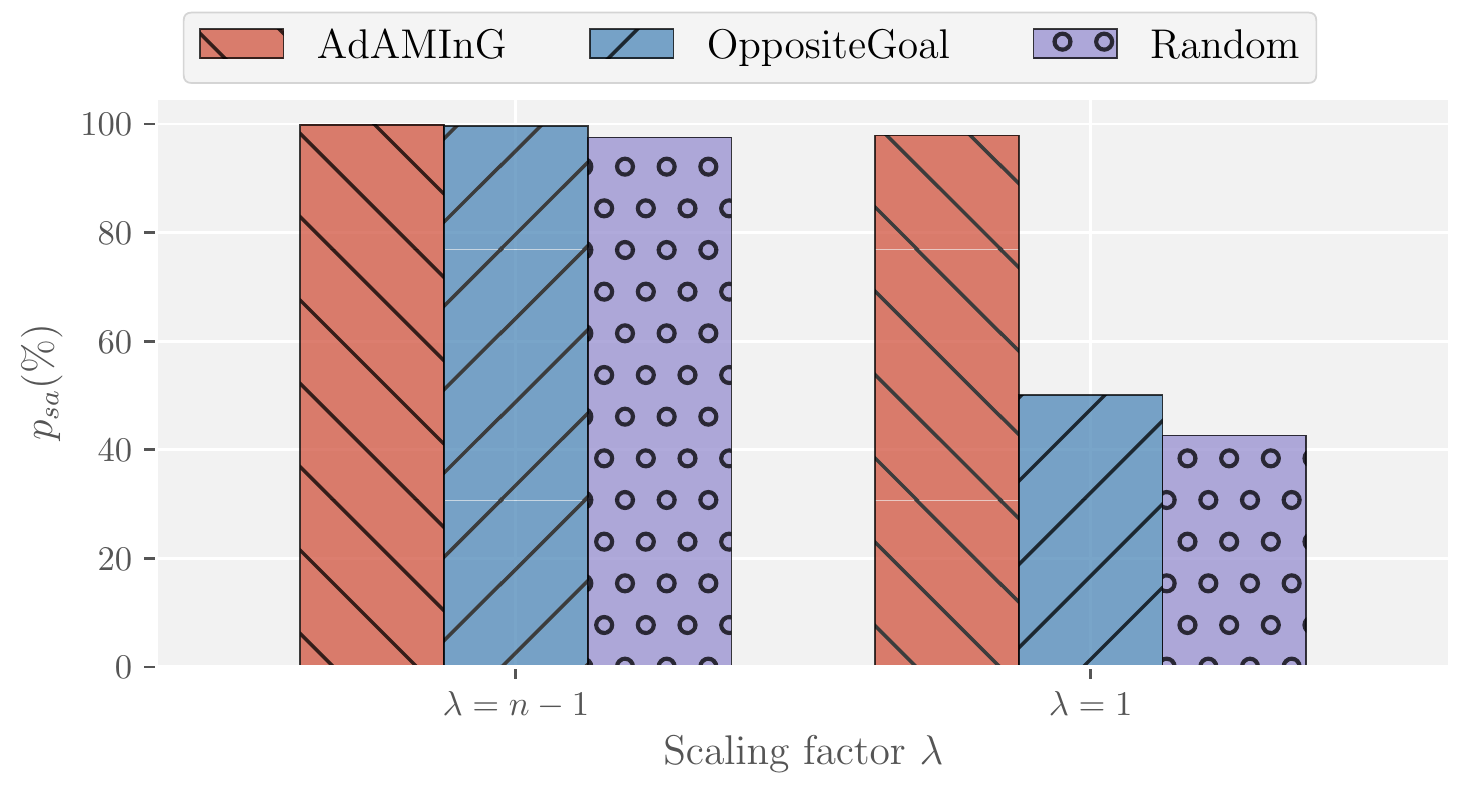}
  \caption{[GridWorld] Probability of successful attack $p_{sa}(\%)$ under different attack models. The greater the $p_{sa}$ the better the performance of the adversary.}
  \label{fig:performance_fedrl}
\end{figure}

\begin{figure}[t]
\centering
  \includegraphics[width=\linewidth]{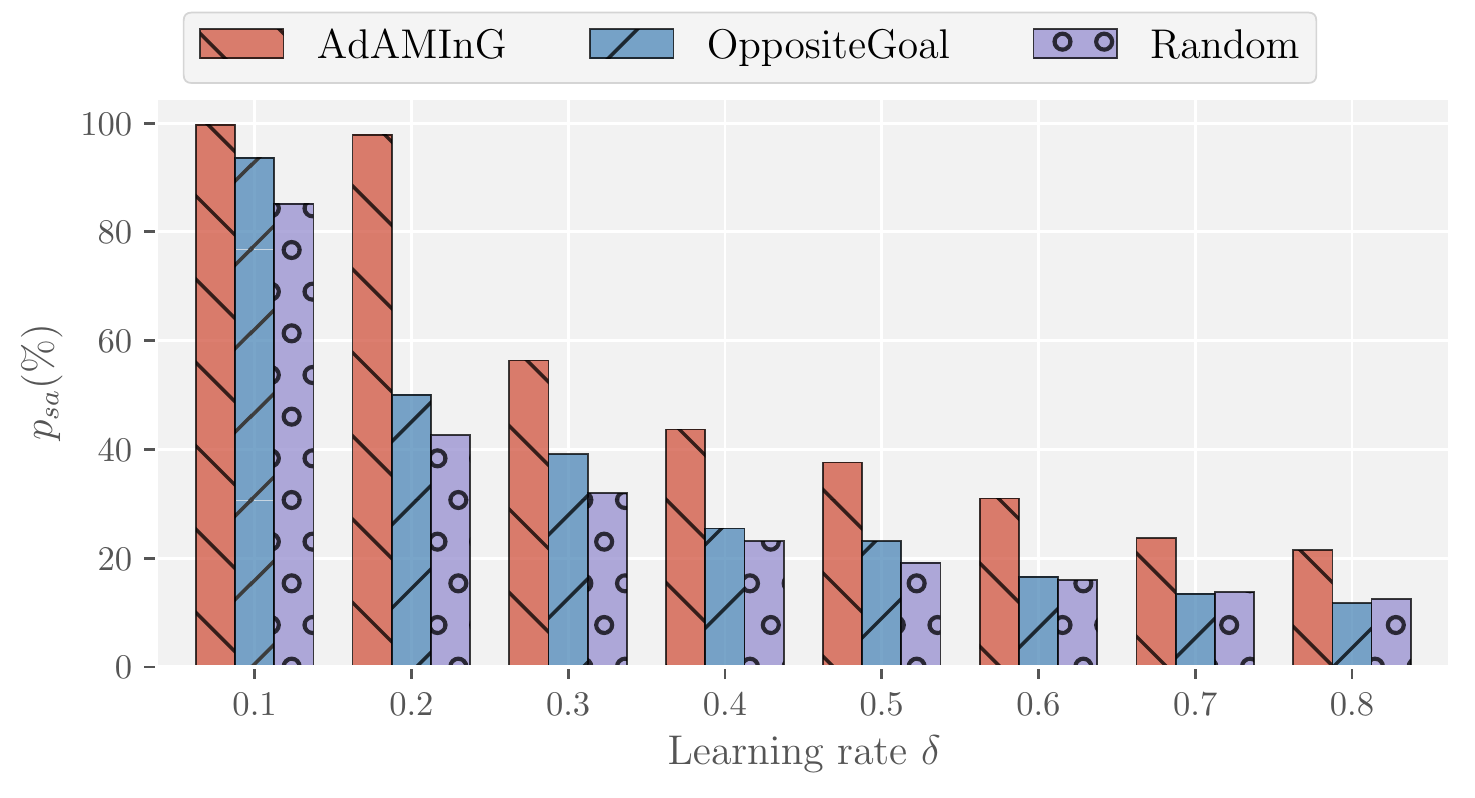}
  \caption{[GridWorld] Effect of learning rate ($\delta$) on the performance of attack methods with $\lambda=1$ and $n=12$.}
  \label{fig:vary_lr_bar}
\end{figure}

\begin{figure}[t]
\centering
  \includegraphics[width=\linewidth]{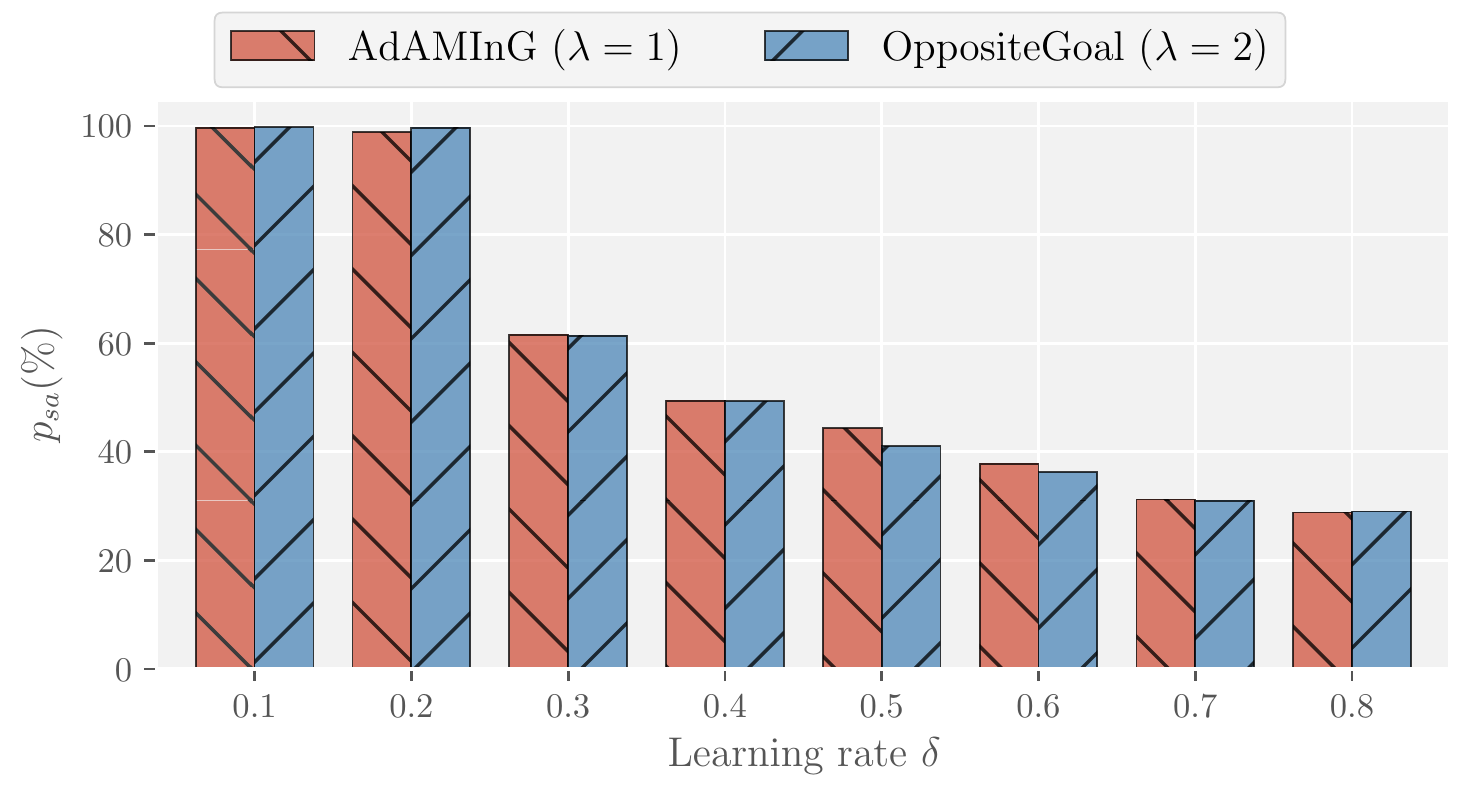}
  \caption{[GridWorld] Comparing attack performance for $n=12$ between \textit{AdAMInG} with $\lambda=1$ and OppositeGoal with $\lambda=2$.}
  \label{fig:vary_lr_bar_multiadv}
\end{figure}

\begin{figure}[t]
\centering
  \includegraphics[width=\linewidth]{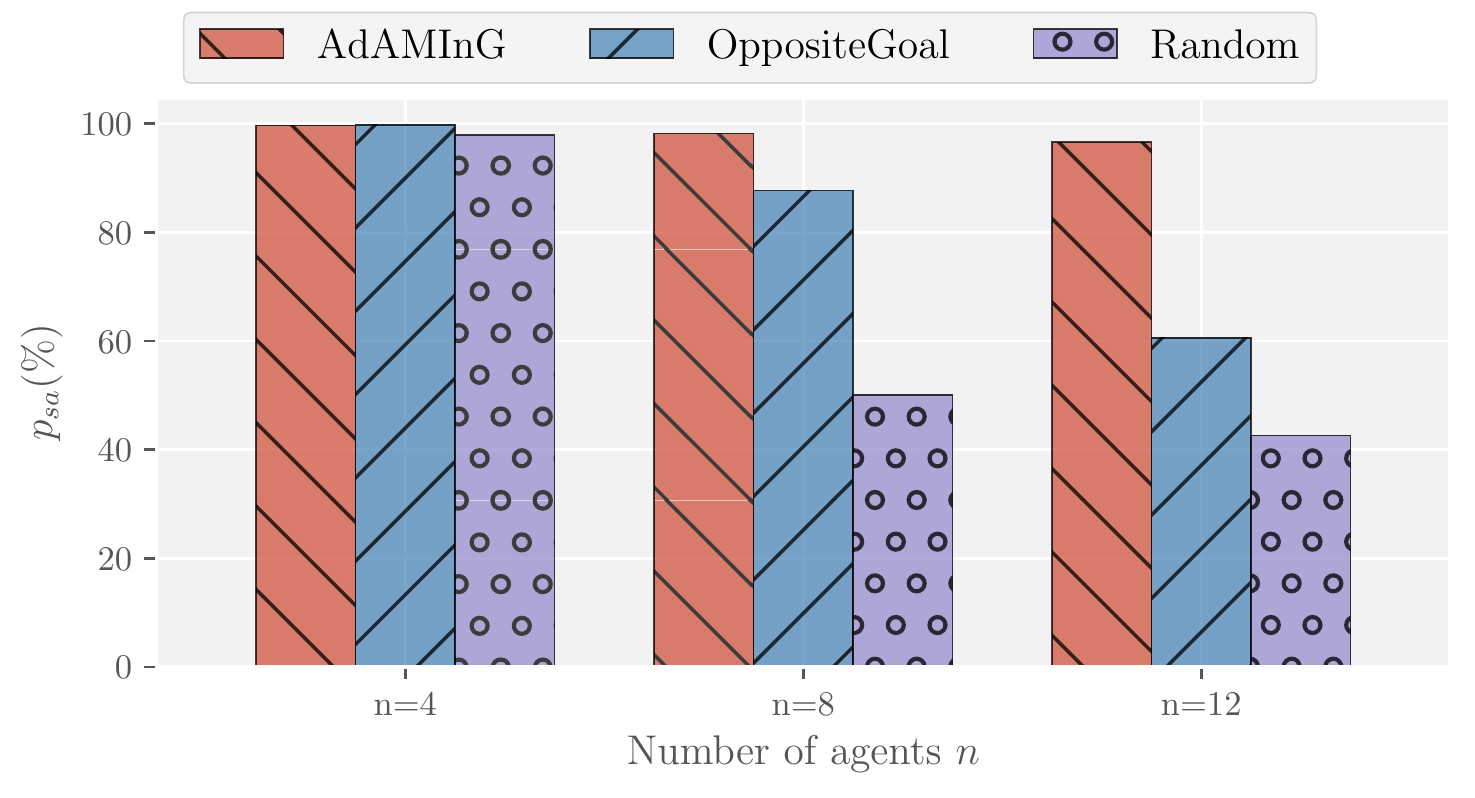}
  \caption{[GridWorld] Effect of number of agents ($n$) on the performance of attack methods with $\lambda=1$ and $\delta=0.2$}
  \label{fig:vary_n_theta_std}
\end{figure}

\begin{figure}[t]
  \includegraphics[width=\linewidth]{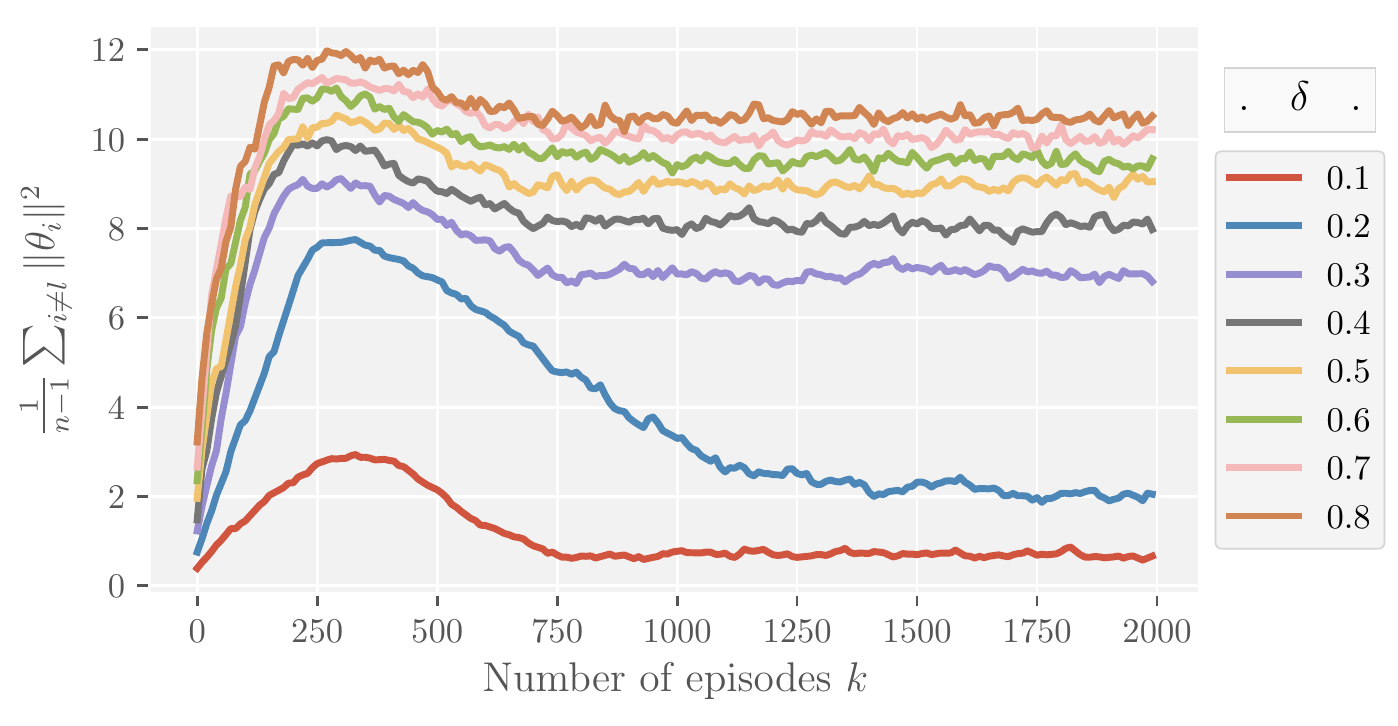}
  \caption{[GridWorld] Based on the learning rate, the consensus gets converged to an intermediate value}
  \label{fig:theta_norm_adam_lr}
\end{figure}

\begin{figure}[t]
  \includegraphics[width=\linewidth]{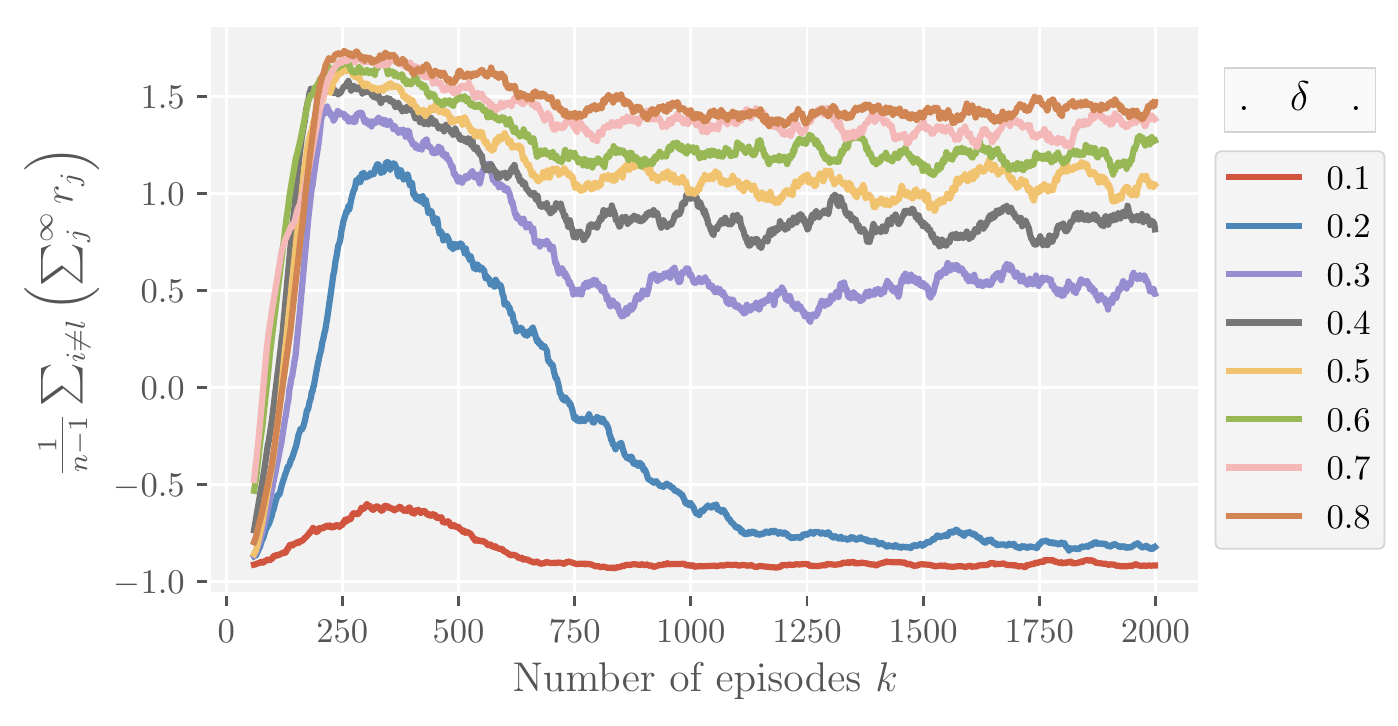}
  \caption{[GridWorld] Cumulative return (moving average of 60) for different learning rate ($\delta$) and $n=12$}
  \label{fig:vary_lr_return}
\end{figure}

\begin{figure}[t]
  \includegraphics[width=\linewidth]{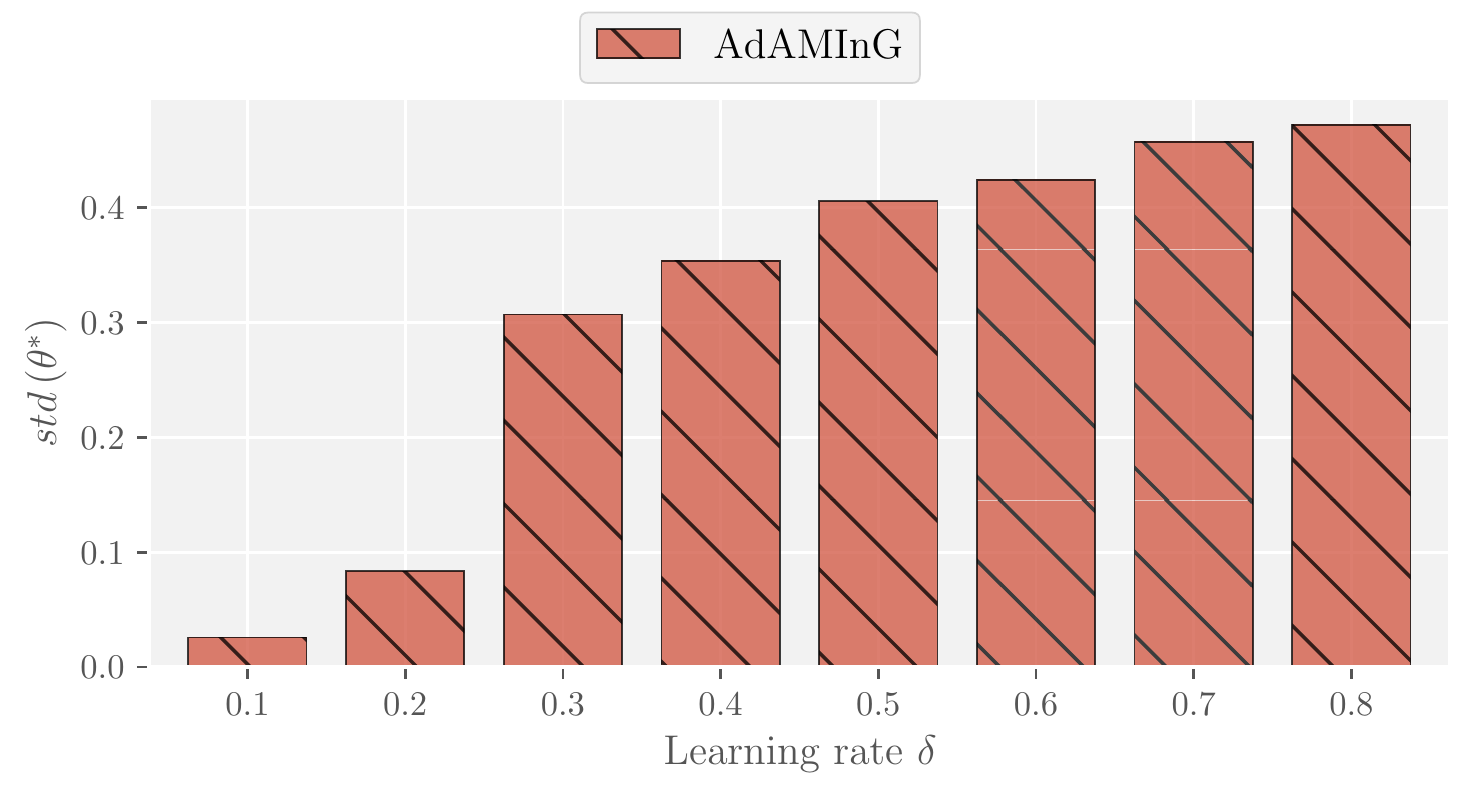}
  \caption{[GridWorld] Standard deviation of the consensus policy parameter }
  \label{fig:vary_lr_theta_std}
\end{figure}

\noindent \textbf{Effect of Adversaries}
We begin the experimentation by analyzing the effect of the common attack models mentioned in Sec. \ref{sec: common_attack_models}.
Fig. \ref{fig:performance_fedrl} reports the $p_{sa}$ for the three attack methods with the scaling factor of $n-1$ and $1$ (and a learning rate $\sigma=0.2$). With the optimal scaling factor of $n-1$, it can be seen that all the three attack methods were able to achieve a good enough attack ($p_{sa}>96\%$). For a scaling factor of 1, however, only \textit{AdAMInG} attack method was able to achieve a successful attack ($p_{sa}=98\%$). Both the random policy attack and OppositeGoal attack were only half as good as the \textit{AdAMInG} attack method with OppositeGoal being only slightly better than a random policy attack.

As mentioned in section \ref{sec:parameters}, for a scaling factor of 1, the performance of the attack method depends on the learning rate ($\delta$) and the number of non-adversarial agents ($n-|\mathcal{L}|$). Fig. \ref{fig:vary_lr_bar} reports $p_{sa}$ of the attack methods with varying learning rates. It can be seen that the greater the learning rate, the poorer the performance of the attack method in terms of $p_{sa}$. For a higher learning rate, the local update for each agent's policy parameter has more effect than the update of the server carried out with the adversary, and hence poorer the performance of the attack. Another thing to observe is that as the learning rate increases the relative performance of the OppositeGoal attack compared to the Random policy attack becomes poorer even becoming worse than the Random policy attack. The reason behind this is that the observable states across environments are non-overlapping. The environment available to the adversary for devising OppositeGoal attack from might not have access to the states observable in other environments. Hence the OppositeGoal policy attack can not modify the policy parameter related to those states. OppositeGoal policy attack method either require a large scaling factor or more than one adversary to attack the MT-FedRL with performance similar to \textit{AdAMInG} with single-adversary and unity scaling factor $\lambda$. This can be seen in Fig. \ref{fig:vary_lr_bar_multiadv}.

A similar trend can be observed with varying the number of non-adversarial agents. It can be seen in Fig. \ref{fig:vary_n_theta_std} that for a smaller number of non-adversarial agents (equivalently smaller number of total agents if the number of the adversarial agents is fixed), it is easier for the adversary to attack with a high $p_{sa}$. The reason behind this is that the local update in Eq. \ref{eq:compromise} is proportional to the number of non-adversarial agents. With a smaller number of non-adversarial agents, the local update is smaller compared to the update by the adversary. Among the three attack methods, \textit{AdAMInG} is the most resilient to these two parameters ($\lambda, n$), hence making it a better choice for an adversarial attack in an MT-FedRL setting.

\noindent \textbf{Analyzing \textit{AdAMInG} Attack}:
We carry out a detailed analysis of the \textit{AdAMInG} attack method. The smoothing average (Eq. \ref{eq:compromise}) in the presence of an adversary carries out two updates - the local update which moves the policy parameter in a direction to maximize the collective goal, and the adversarial update which tries to move the policy parameter away from the consensus. When the training begins, the initial set of policy parameters $\theta_i$ is farther away from the consensus $\theta^*$. Gradient descent finds a direction from the current set of policy parameters to the consensus. This direction has a higher magnitude when the distance between the current policy parameter and the consensus is high. As the system learns, the current policy parameter gets closer to the consensus, and hence the magnitude of the direction of update decreases. So even if we have a static learning rate $\delta$, the magnitude of local update $\delta_j \nabla_{\theta_j}V_{j}^{\pi_{\theta_j}}(\rho_{j})$ in Eq. \ref{eq:compromise} will, in general, decrease as the system successfully learns.
There will be a point in training where the local update will become equal but opposite to the update being carried out by the \textit{AdAMInG} adversary. From that point onwards, the current policy parameter won't change much. This can be seen in Fig. \ref{fig:theta_norm_adam_lr}. The greater the learning rate $\delta$, the earlier in training we will get to the equilibrium point, and hence poorer the attack performance which can be seen in terms of the achieved discounted return in Fig. \ref{fig:vary_lr_return}. A greater standard deviation of the consensus policy parameter indicates a better differentiation between good and bad actions for a given state. Fig. \ref{fig:vary_lr_theta_std} plots the standard deviation of the consensus policy parameter for different learning rates $\delta$. It can be seen that for higher learning rates, the consensus has a higher standard deviation hence being able to perform better than the consensus achieved under lower learning rates.

\begin{figure}[t]
  \includegraphics[width=\linewidth]{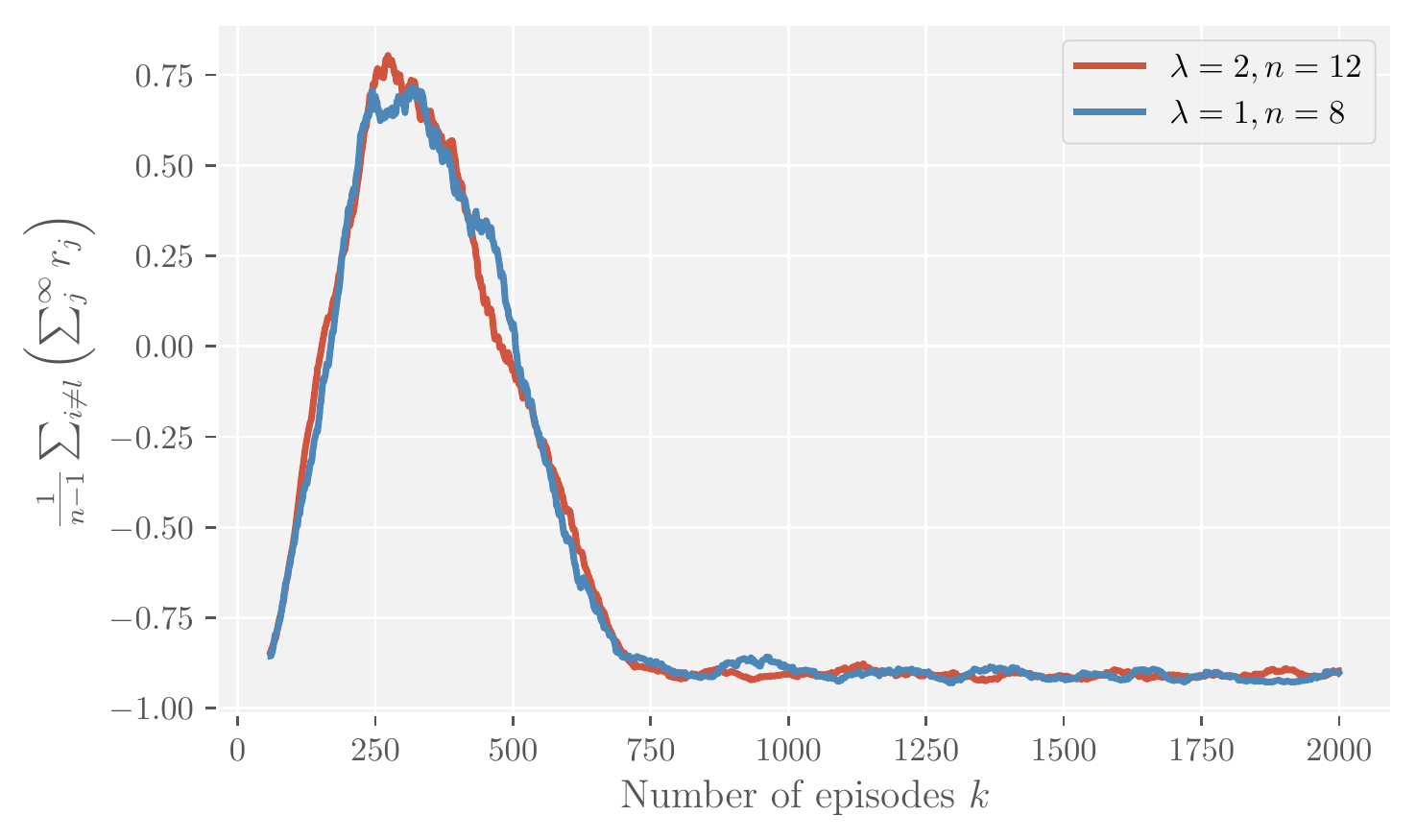}
  \caption{[GridWorld] Relationship between $\lambda$ and $n$ for same \textit{AdAMInG} attack performance. $\left(\lambda=1, n=8\right)$ and $\left(\lambda=2, n=12\right)$ follows the same discounted return across episodes which is in accordance with Eq. \ref{eq:lambda_n_relation}}
  \label{fig:relation_lambda_n_theta}
\end{figure}

We also compare the performance of the \textit{AdAMInG} attack in relation to the scaling factor $\lambda$ and the number of agents $n$. According to Eq. \ref{eq:lambda_n_relation} an increase in the number of agents can be compensated by increasing the scaling factor $\lambda$ to achieve the same attacking performance. We analyse the \textit{AdAMInG} attack for the following two configurations: $\left(\lambda=1, n=8\right)$ and $\left(\lambda=2, n=12\right)$. Table \ref{table:lambda_n} reports the $p_{sa}$ and the standard deviation of the consensus policy parameter $\theta^*$. It can be seen that both the configurations generate similar numbers. The same trend can be observed temporally, in Fig. \ref{fig:relation_lambda_n_theta}, for the achieved discounted return during each episode in training.

\begin{table}
\centering
\begin{tabular}{cccc}
\toprule
\textbf{Configuration}                & \textbf{Learning rate} $\delta$           & $p_{sa}\%$   &$std$       \\ \toprule
$\lambda=1, n=8$                & 0.2                & 99.75\%      & 0.036      \\  
$\lambda=2, n=12$               & 0.2                & 99.49\%      & 0.031      \\ \bottomrule
\end{tabular}
\caption{[GridWorld] Relationship between $\lambda$ and $n$ for same attack performance with \textit{AdAMInG}}
\label{table:lambda_n}
\end{table}

\begin{figure}[t]
\centering
  \includegraphics[width=\linewidth]{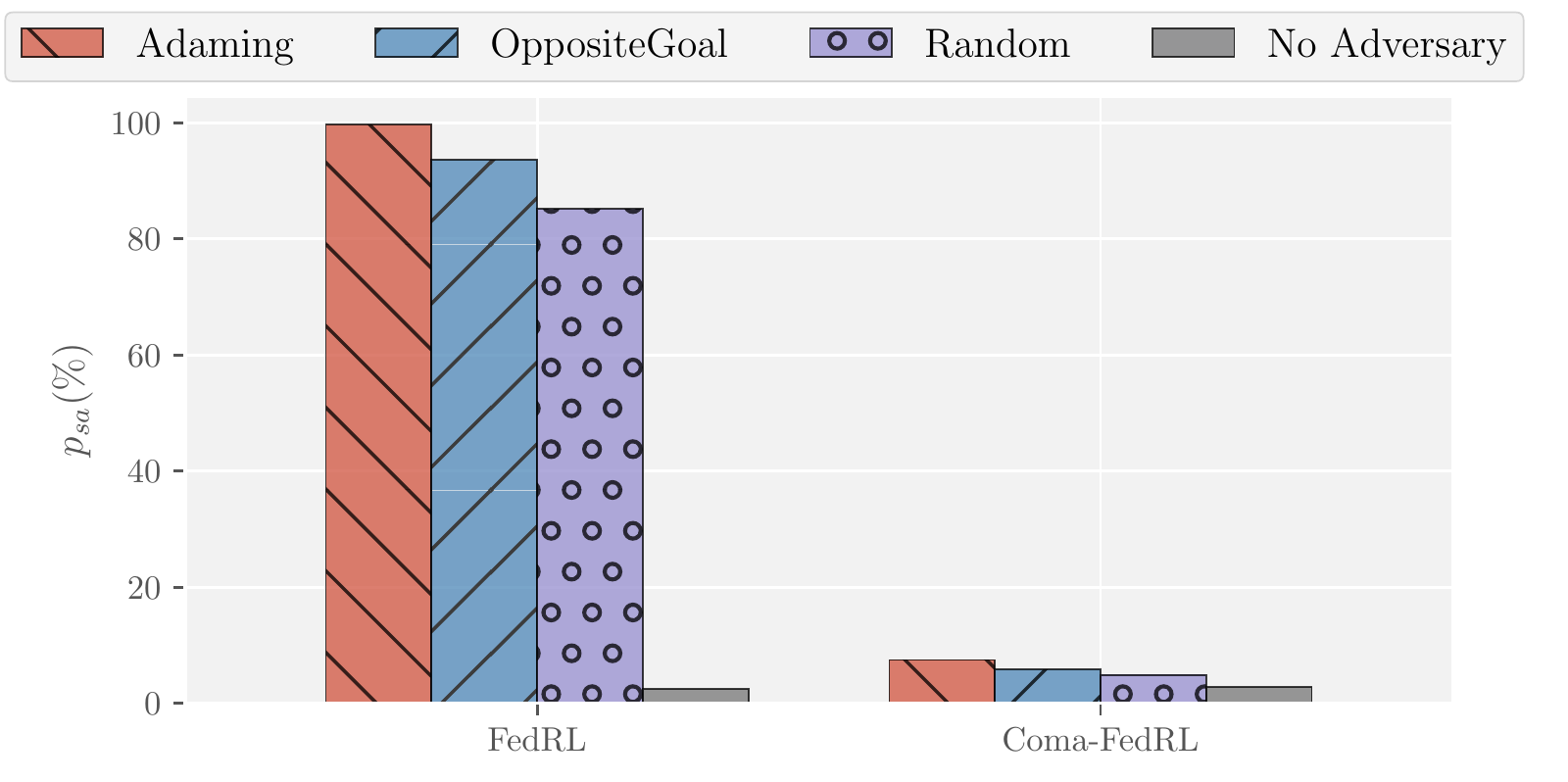}
  \caption{[GridWorld] Comparison of probability of successful attack $p_{sa}(\%)$ under different attack models for \texttt{FedRL} and \texttt{ComA-FedRL}. The effect of adversarial agent is greatly reduced with \texttt{ComA-FedRL}.}
  \label{fig:performance_comafedrl}
\end{figure}

\begin{figure}[t]
  \includegraphics[width=\linewidth]{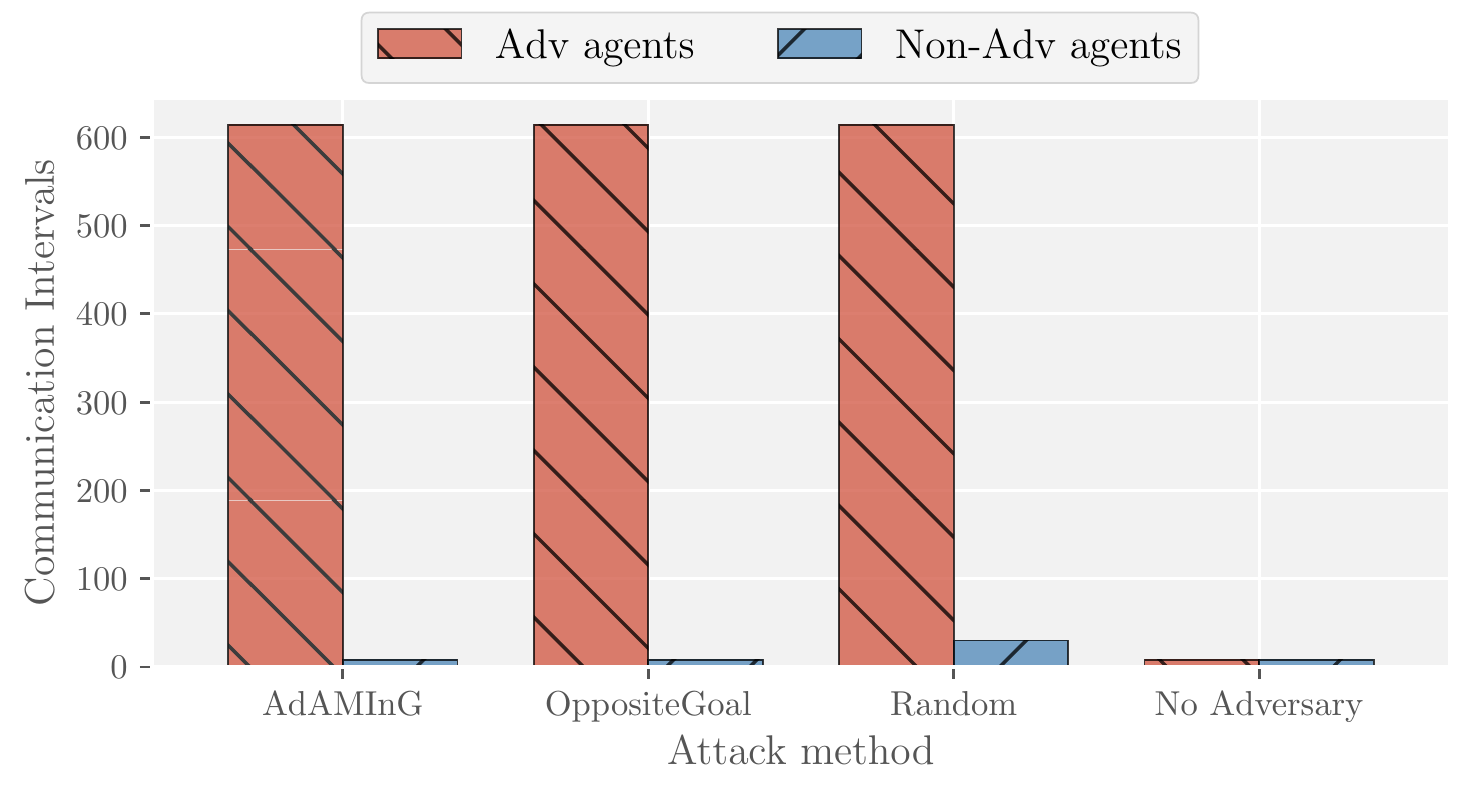}
  \caption{[GridWorld] Average communication intervals for adversarial and non adversarial agents in \texttt{ComA-FedRL}}
  \label{fig:comm_comafedrl}
\end{figure}

\noindent \textbf{Resolving adversaries:}
We implement the N-agent single-adversary MT-FedRL problem using \texttt{ComA-FedRL} to address the high $p_{sa}$ of the conventional FedRL algorithm. Fig. \ref{fig:performance_comafedrl} compares the performance of FedRL and \texttt{ComA-FedRL} for different attack methods. By assigning a higher communication interval to the probable adversary, \texttt{ComA-FedRL} was able to decrease the probability of successful attack $p_{sa}$ in the presence of adversary to as low as $<10\%$. The mean communication interval for adversarial and non-adversarial agents is plotted in Fig. \ref{fig:comm_comafedrl}. It can be seen that Random policy attack has a slightly higher communication interval. The reason behind this is one of the non-adversarial agents was incorrectly marked as a probable adversarial agent at the beginning of training, but later that was self-corrected to a \textit{possibly-non-adversarial} agent.

\begin{figure}[t]
\centering
  \includegraphics[width=0.85\linewidth]{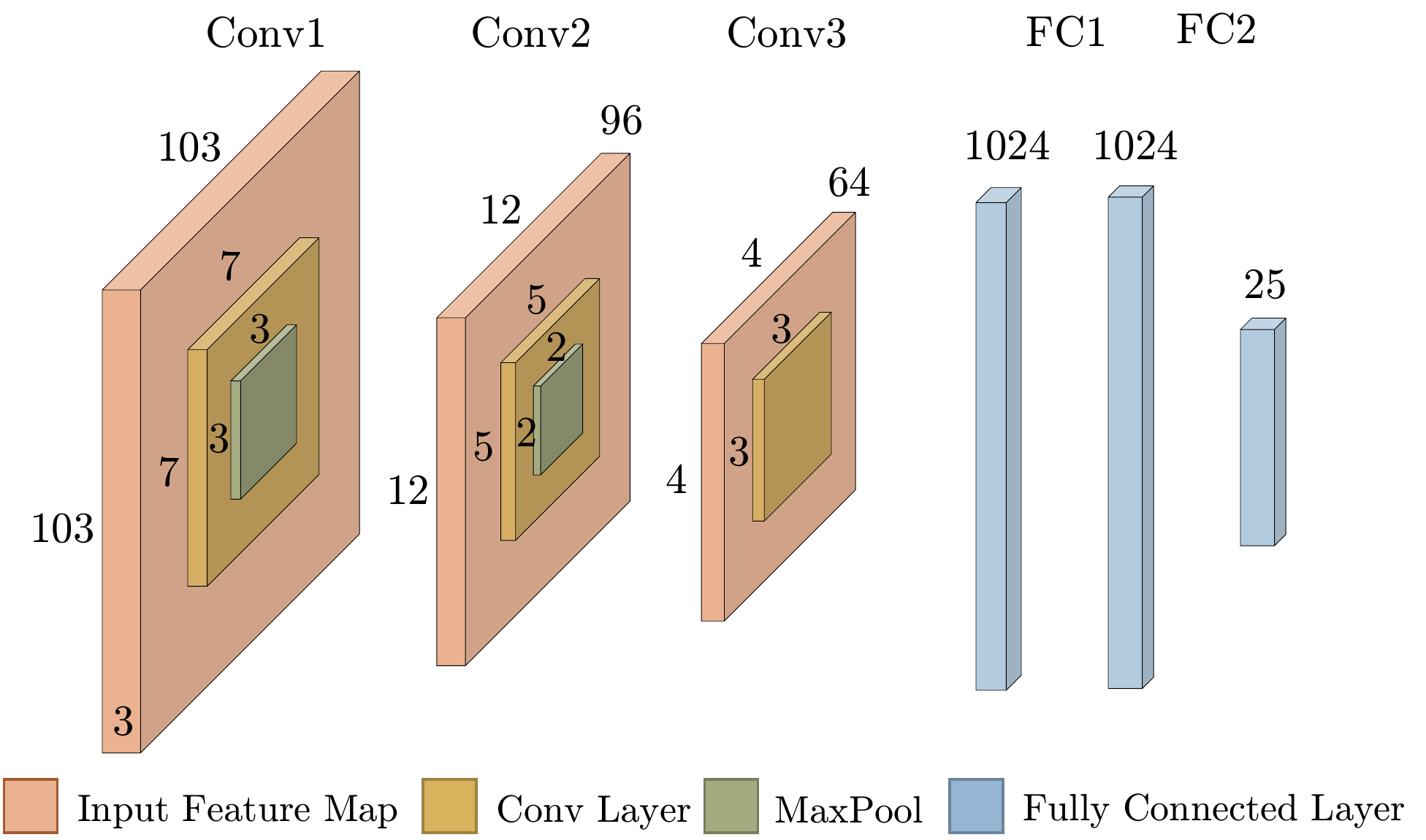}
  \caption{[AutoNav] C3F2 neural network used to map states to action probabilities}
  \label{fig:C3F2}
\end{figure}

\begin{figure*}[t]
\centering
  \includegraphics[width=\linewidth]{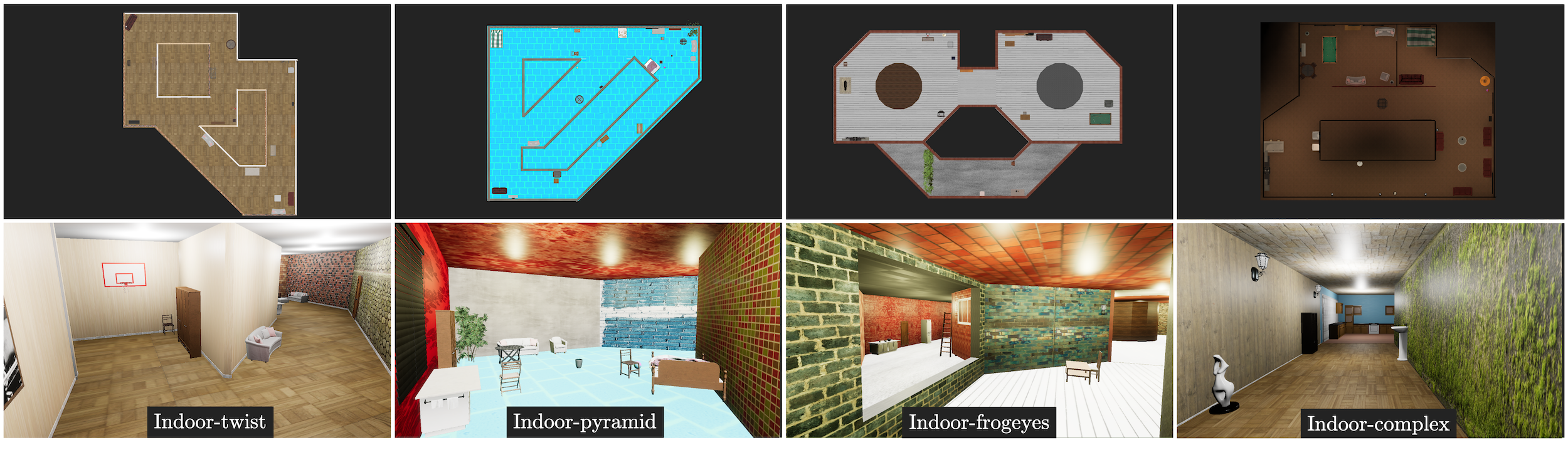}
  \caption{[AutoNav] Floor plan and screenshot of the four 3-D environments used}
  \label{fig:pedra_env}
\end{figure*}

\subsection{AutoNav - NN based RL}
\noindent \textbf{Problem Description:}
We also experiment on a more complex problem of drone autonomous navigation in 3D realistic environments. We use PEDRA \cite{anwar2020autonomous} as the drone navigation platform. The drone is initialized at a starting point and is required to navigate across the hallways of the environments. There is no goal position, and the drone is required to fly avoiding the obstacles as long as it can. At each iteration $t$, the drone captures an RGB monocular image from the front-facing camera which is taken as the state $s_t \in \mathbb{R}^{(320\times 180 \times 3)}$ of the RL problem. Based on the state $s_t$, the drone takes an action $a_t\in \mathcal{A}$. We consider a perception based probabilistic action space with 25 actions ($|\mathcal{A}|=25$). A depth-based reward function is used to encourage the drone to stay away from the obstacles. We use neural network-based function approximation to estimate the action probabilities based on states. The C3F2 network used is shown in Fig. \ref{fig:C3F2}. We consider 4 indoor environments (indoor-twist, indoor-frogeyes, indoor-pyramid, and indoor-complex) hence we have $n=4$. These environments can be seen in Fig. \ref{fig:pedra_env}.

The effectiveness of MT-FedRL-achieved unified policy is quantified by Mean Safe Flight (MSF) defined as
\begin{align*}
    MSF = \frac{1}{n-1}\mathbb{E} \left[ \sum_{i\neq l} d_i\right]
\end{align*}
where $d_i$ is the distance traveled by the agent in the environment $i$ before crashing.
In this 4-agent MT-FedRL system, the agent in the environment indoor-complex is assigned the adversarial role ($l=3$). The goal for the adversarial agent is to decrease this MSF. We will characterize the performance of the adversarial attack by the probability of successful attack $p_{sa}$ given by
\begin{align*}
    p_{sa} = 1 -  \frac{MSF_{adv}}{MSF_{no-adv}}
\end{align*}
where $MSF_{adv}$ is the mean safe flight of the MT-FedRL system in the presence of the adversary, while $MSF_{no-adv}$ is the mean safe flight of the MT-FedRL system in the absence of the adversary. The greater the $p_{sa}$ the better the attack method in achieving its goal.

\begin{figure}[t]
  \includegraphics[width=\linewidth]{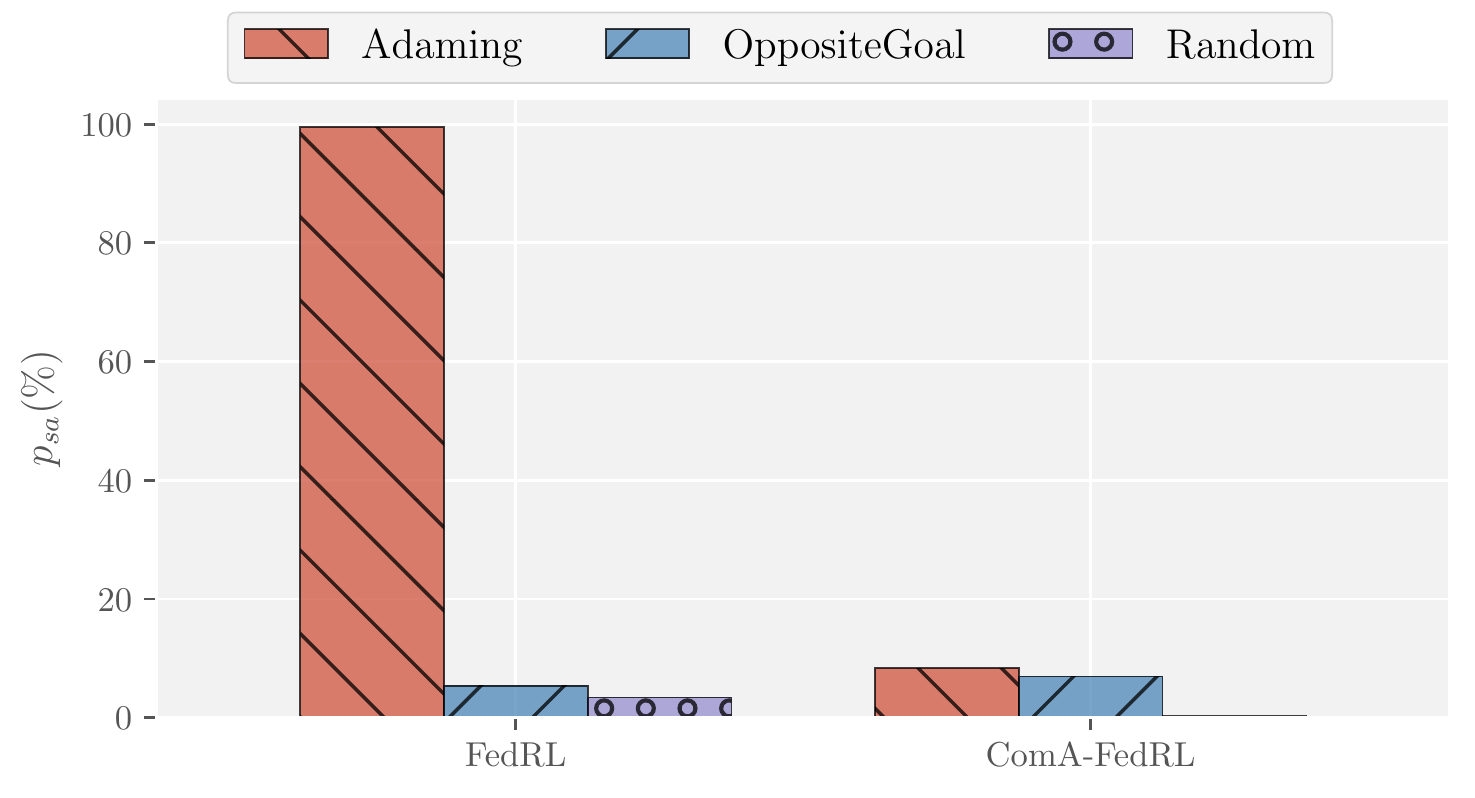}
  \caption{[AutoNav]  Comparison of probability of successful attack $p_{sa}(\%)$ under different attack models for \texttt{FedRL} and \texttt{ComA-FedRL}. The effect of adversarial agent is greatly reduced with \texttt{ComA-FedRL}.}
  \label{fig:autonav_performance}
\end{figure}

\noindent \textbf{Effect of Adversaries:} For each experiment, the MT-FedRL problem is trained for 4000 episodes using the REINFORCE algorithm with a learning rate of 1e-4 and $\gamma=0.99$. Training hyper-parameters are listed in the appendix section \ref{sec:training_details} in detail. Table \ref{tab:autonav} reports the MSF achieved by the AutoNav problem for various attack methods. It can be seen that except for the \textit{AdAMInG} attack, the rest of the attack methods achieve MSF comparable to the one achieved in the absence of an adversary ($\sim 1000m$).
Fig. \ref{fig:autonav_performance} plots the $p_{sa}$ for the different attack methods. It can be seen that \textit{AdAMInG} achieves a $p_{sa}$ of $\sim 99.5\%$ while all the other attack methods achieve a $p_{sa}$ of $< 6\%$. The trend is similar to what was observed in the GridWorld task

\noindent \textbf{Resolving Adversaries:} We implement the N-agent single-adversary MT-FedRL problem using \texttt{ComA-FedRL} to address the low MSF of FedRL. The results are reported in Table \ref{tab:autonav}. It can be seen that the decrease in MSF due to adversary was recovered using \texttt{ComA-FedRL}. Fig. \ref{fig:autonav_performance} plots the $p_{sa}$ for various attack methods with \texttt{ComA-FedRL} and compares it with \texttt{FedRL}. It can be see that with \texttt{ComA-FedRL} we have $p_{sa} < 10\%$. Hence \texttt{ComA-FedRL} was able to address the issue of adversaries in a MT-FedRL problem.

\begin{table}[t]
\begin{tabular}{ccccc}
\toprule
                    & \textbf{\textit{AdAMInG}}  & \textbf{Opposite Goal}    & \textbf{Random}   & \textbf{No Adv} \\ \toprule
\textbf{FedRL}      & 6                 & 1076                      & 1098              &   1137     \\
\textbf{ComA-FedRL} & 1042              & 1028                      & 1134              & 1156   \\ \toprule
\end{tabular}
\caption{[AutoNav] MSF (m) for different attack methods}
\label{tab:autonav}
\end{table}

\section{Conclusion}
\label{sec:conclusion}
In this paper we analyse Multi-task Federated Reinforcement Learning algorithm with an adversarial perspective.
We analyze the attacking performance of some general attack methods and propose an adaptive attack method \textit{AdAMInG} that devises an attack taking into account the aggregation operator of federated RL. The \textit{AdAMinG} attack method is formulated and its effectiveness is studied. Furthermore, to address the issue of adversaries in MT-FedRL problem, we propose a communication adaptive modification to conventional federated RL algorithm, \texttt{ComA-FedRL}, that varies the communication frequency for the agents based on their probability of being an adversary. Results on the problems of GridWorld (maze solving) and AutoNav (drone autonomous navigation) show that the \textit{AdAMInG} attack method outperforms other attack methods almost every time. Moreover, \texttt{ComA-FedRL} was able to recover from the adversarial attack resulting in near-optimal policies.
\section{Acknowledgements} 
This work was supported in part by C-BRIC, one of six centers in JUMP, a Semiconductor Research Corporation (SRC) program sponsored by DARPA.

\bibliographystyle{ieeetran}
\bibliography{IEEEabrv,main}

\begin{IEEEbiography}[{\includegraphics[width=1in,height=1.25in,clip,keepaspectratio]{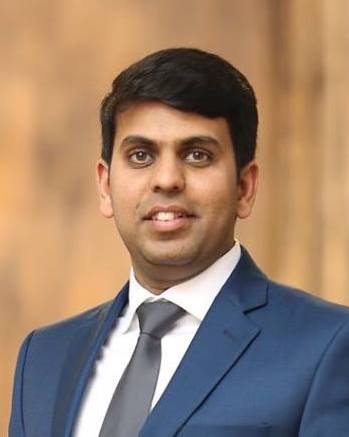}}]{Aqeel Anwar}
received his Bachelor's degree in Electrical Engineering from the University of Engineering and Technology (UET), Lahore, Pakistan, and Masters degree in Electrical and Computer Engineering from Georgia Institute of Technology, Atlanta, GA, USA in 2012 and 2017 respectively. Currently, he is pursuing his Ph.D. in Electrical and Computer Engineering from the Georgia Institute of Technology under the supervision of Dr. Arijit Raychowdhury.
His research interests lie at the junction of machine learning and hardware design. He is working towards shifting Machine Learning (ML) from cloud to edge nodes by improving the energy efficiency of current state-of-the-art ML algorithms and designing efficient DNN accelerators.
\end{IEEEbiography}

\begin{IEEEbiography}[{\includegraphics[width=1in,height=1.25in,clip,keepaspectratio]{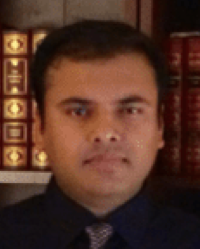}}]{Arijit Raychowdhury}
(Senior Member, IEEE) received the Ph.D. degree in electrical and computer engineering from Purdue University, West Lafayette, IN, USA, in 2007. His industry experience includes five years as a Staff Scientist with the Circuits Research Laboratory, Intel Corporation, Portland, OR, USA, and a year as an Analog Circuit Researcher with Texas Instruments Inc., Bengaluru, India. He joined the Georgia Institute of Technology, Atlanta, GA, USA, in 2013, where he is currently an Associate Professor with the School of Electrical and Computer Engineering and also holds an ON Semiconductor Junior Professorship. He holds more than 25 U.S. and international patents and has published over 100 papers in journals and refereed conferences. His research interests include low-power digital- and mixed-signal circuit design, device–circuit interactions, and novel computing models and hardware realizations. Dr. Raychowdhury was a recipient of the Dimitris N. Chorafas Award for Outstanding Doctoral Research in 2007, the Intel Labs Technical Contribution Award in 2011, the Best Thesis Award from the College of Engineering, Purdue University, in 2007, the Intel Early Faculty Award in 2015, the NSF CISE Research Initiation Initiative Award (CRII) in 2015, and multiple best paper awards and fellowships.
\end{IEEEbiography}

\newpage 
\appendix
\subsection{Training details} \label{sec:training_details}
Policy gradient methods for RL is used to train both the GridWorld and AutoNav RL problems. For \texttt{ComA-FedRL}, we use a base communication $base\_comm$. In the pre-train phase, the communication interval for each agent is assigned this base communication i.e.
\begin{align*}
    comm[i] = base\_comm \quad \forall i \in \{0, n-1\}
\end{align*}
This means that in the pre-train phase, the agents learn only on local data, and after every $base\_comm$ number of episodes, the locally learned policies are shared with the server for cross-evaluation. This cross-evaluation runs $n$ policies, each on a randomly selected environment and the cumulative reward is recorded. We also take into account the fact that the adversarial agent can present a secondary attack in terms of faking the cumulative reward that it return when evaluating a policy. In the ComA-FedRL implementation, we assume that the adversarial agent returns a cumulative reward of $-1$, meaning that it fakes the policy being evaluated as adversarial.

At the end of the pre-train phase, the cross evaluated rewards are used to assign communication intervals to all the agents. There are various choices for the selection of this mapping. The underlying goal is to assign a higher communication interval for agents whose policy performs poorly when cross-evaluated and vice versa. We use the mapping shown in Alg. \ref{Alg:comm_int}. A reward threshold $r_{th}$ is used to assign agents different communication intervals. If the cumulative reward of a policy in an environment is below $r_{th}$, it is assigned a high communication interval of $high\_comm$ episodes (marked as a possible adversary), otherwise it is assigned a low communication interval of $low\_comm$ episodes (marked as a possible non-adversary). The assigned communication interval also depends on the one-step history of communication intervals. If an agent was previously assigned a higher communication interval and is again marked as a possible adversary, the communication interval assigned to such an agent is doubled.
The complete list of hyperparameters used for GridWorld and AutoNav can be seen in Table \ref{tab:hyperparameters}.

\begin{algorithm}[t]
\SetAlgoLined

\SetKwFunction{FMain}{UpdateCommInt}
\SetKwProg{Pn}{Function}{:}{}
\Pn{\FMain{$r_{m\times n}$, $comm$}}{
Initialize $low\_comm,~high\_comm,~r_{th}$

\For{each agent i}{
Average the rewards across episodes
\begin{align*}
    r_{avg} \leftarrow \frac{1}{m}\sum_{j=0}^{m-1} r[:, i]
\end{align*}
    \If{$r_{avg} \geq r_{th}$}{
    $comm[i] = low\_comm$
    }
    \ElseIf{$r_{avg} < r_{th}$}
    {
        \If{$comm[i]\neq low\_comm$}
        {   
            $comm[i] = 2*comm[i]$
        }
        \Else
        {
            $comm[i] = high\_comm$
        }
        
    }
    
    }
}
\KwRet  $comm$

\caption{Update Communication Intervals}
\label{Alg:comm_int}
\end{algorithm}

\begin{table}[b]
\centering
\begin{tabular}{lcc}
\bottomrule
\multicolumn{1}{c}{\textbf{HyperParameter}} & \textbf{GridWorld} & \textbf{AutoNav} \\ \toprule
Functional Mapping                          & Tabular          & Neural Network    \\
Number of agents                            & 4, 8, 12         & 4                 \\
Algorithm                                   & REINFORCE        & REINFORCE          \\
Max Episodes                                & 1000             & 4000               \\
Gamma                                       & 0.95             & 0.99               \\
Learning rate                               & Variable         & 1e-4           \\
$base\_comm$                                & 8                & 8                 \\
$wait\_train$                               & 600              & 1000           \\
Gradient clipping norm                      & None             & 0.1               \\
Optimizer type                              & ADAM             & ADAM              \\
Entropy Regularizer Scalar                  & None             & 0.5               \\
Training Workstation                        & GTX1080          & GTX1080            \\
Training Time                               & 9 hours          & 35 hours           \\ 

\toprule
\end{tabular}
\caption{Training hyper-parameters for GridWorld and AutoNav}
\label{tab:hyperparameters}
\end{table}

\end{document}